\documentclass[10pt, twocolumn]{article}
\usepackage[
  a4paper,
  top=20mm, bottom=20mm,
  left=15mm,  right=15mm,
  columnsep=6mm,
  headheight=14pt
]{geometry}

\usepackage[T1]{fontenc}
\usepackage[utf8]{inputenc}
\usepackage{lmodern}
\usepackage{microtype}

\usepackage{amsmath, amssymb, amsfonts, amsthm}
\usepackage{bm}

\usepackage{graphicx}
\usepackage{subcaption}
\usepackage{booktabs}
\usepackage{multirow}
\usepackage{array}
\usepackage{float}
\usepackage{placeins}
\usepackage{tikz}
\usetikzlibrary{positioning,arrows.meta,calc,decorations.pathreplacing}
\usepackage{enumitem}
\usepackage{xcolor}
\usepackage{xspace}
\usepackage{algorithm}
\usepackage{algpseudocode}

\definecolor{ttorange}{HTML}{EB684C}
\definecolor{sonatablue}{HTML}{35636E}
\definecolor{sagegreen}{HTML}{8B9D77}

\usepackage[
  colorlinks=true,
  linkcolor=ttorange,
  citecolor=ttorange,
  urlcolor=ttorange
]{hyperref}
\usepackage{cleveref}

\usepackage[numbers, sort&compress]{natbib}

\usepackage{fancyhdr}
\usepackage{titlesec}
\titleformat{\section}{\large\bfseries}{\thesection}{1em}{}
\titleformat{\subsection}{\normalsize\bfseries}{\thesubsection}{1em}{}
\titleformat{\subsubsection}{\normalsize\itshape}{\thesubsubsection}{1em}{}
\titlespacing*{\section}{0pt}{12pt}{4pt}
\titlespacing*{\subsection}{0pt}{8pt}{2pt}

\usepackage[affil-it]{authblk}

\usepackage{etoolbox}
\newcommand{\papermode}{V1} % Change to Draft, V1, V1.1, etc.

\newcommand{\todo}[1]{%
  \ifdefstring{\papermode}{Draft}{\textcolor{red}{\textbf{[TODO: #1]}}}{}%
}

\newcommand{\inmode}[2]{%
  \ifdefstring{\papermode}{#1}{#2}{}%
}

\newcommand{\sonata}{\textsc{Sonata}\xspace}

\newcommand{\hlines}[1]{%
  \foreach \pct in {0.20, 0.40, 0.60, 0.80}{%
    \draw[white, semithick]
      ($(#1.south west)!\pct!(#1.north west)$) --
      ($(#1.south east)!\pct!(#1.north east)$);%
  }%
}

\title{\textbf{\textsc{Sonata}: A Hybrid World Model for Inertial Kinematics under Clinical Data Scarcity}}

\author[1]{Blaise Delaney}
\author[1]{Salil Patel}
\author[1]{Yuji Xing}
\author[1]{Dominic Dootson}
\author[2]{Karin Sevegnani}
\author[3]{Chrystalina Antoniades}

\affil[1]{TimeTrace Labs}
\affil[2]{NVIDIA}
\affil[3]{University of Oxford}

\begin{document}

\twocolumn[{%
  \maketitle

  \begin{center}
    {\small Correspondence: blaise@timetracelabs.com}
  \end{center}
  \vspace{0.2em}

  \begin{center}
    \begin{minipage}{0.9\textwidth}
      \begin{center}
        \textbf{Abstract}
      \end{center}
      \small
      We introduce \sonata, a compact latent world model for six-axis trunk IMU
      representation learning under clinical data scarcity. Clinical cohorts
      typically comprise tens to hundreds of patients, making web-scale
      masked-reconstruction objectives poorly matched to the problem. \sonata is
      a 3.77\,M-parameter hybrid model, pretrained on a harmonised corpus of
      nine public datasets ({$\sim$}739 subjects, {$\sim$}190k windows) with a
      latent world-model objective that predicts future state rather than
      reconstructing raw sensor traces. In a controlled comparison against a
      matched autoregressive forecasting baseline (MAE) on the same backbone,
      \sonata yields consistently stronger frozen-probe clinical discrimination,
      prospective fall-risk prediction, and cross-cohort transfer across a
      14-arm evaluation suite, while producing higher-rank, more structured
      latent representations. At 3.77\,M parameters the model is compatible
      with on-device wearable inference, offering a step toward general
      kinematic world models for neurological assessment.
    \end{minipage}
  \end{center}
  \vspace{1em}
}]
\thispagestyle{fancy}

% -----------------------------------------------------------------------
% Main sections
% -----------------------------------------------------------------------

% sections/01_intro.tex
% Introduction

\section{Introduction}\label{sec:intro}

In many clinical domains, the early signatures of disease are encoded in continuous physiological time series long before they surface in a clinic: the cardiac trace that first reveals an arrhythmia, the EEG that anticipates a seizure, the respiratory waveform that marks the onset of decline.
In neurology, where dysfunction manifests in the mechanics of movement, this signal is kinematic---and it is accessible, non-invasively, through the body in motion, raising a concrete representation learning problem.

The gait patterns that reveal early Parkinson's disease, or distinguish progressive multiple sclerosis from stable remission, are legible in the raw sensor signal.
They are encoded in the amplitude and timing of each footfall, in the rotational dynamics of the trunk through a turn, in the subtle asymmetries that accumulate over a walk.
Wearable inertial sensors can capture these patterns continuously, outside the clinic, and without interrupting daily life~\citep{Li2025FallDecade,Palmerini2023MobiliseD}.
The biological signal is rich and the measurement problem is tractable.
The constraint is elsewhere.

\begin{figure}[!t]
\centering
\captionsetup{font=small, skip=4pt}
\includegraphics[width=0.93\linewidth]{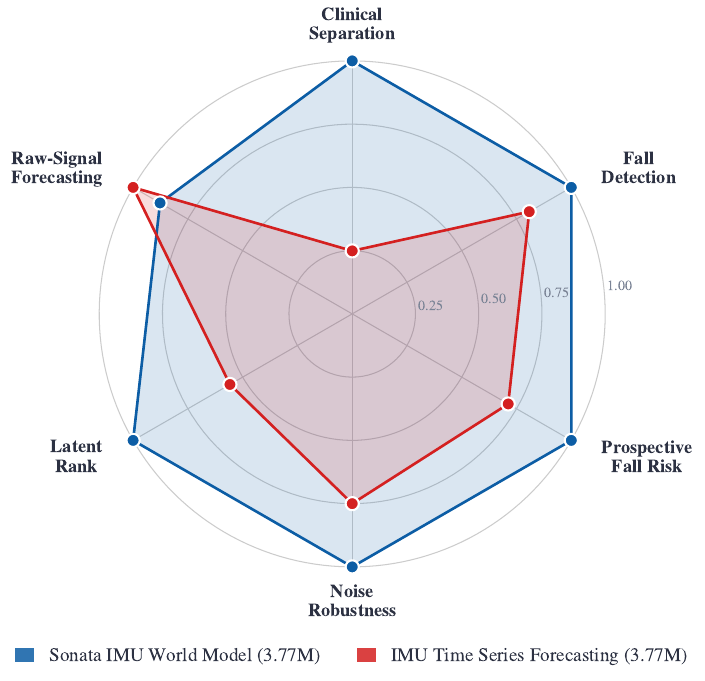}
\caption{Radar summary of representation quality comparing \textbf{\textsc{Sonata}}
against a same-backbone IMU forecasting baseline across six evaluation
domains. Values are normalized per axis; higher is better. \textsc{Sonata} is
stronger on the clinically oriented mix of discrimination, robustness, and
representation-quality probes, while forecasting is stronger on raw-signal
reconstruction. Both variants share the same encoder backbone; legend
parameter counts refer to total model size including the respective
pretraining heads.}
\label{fig:radar_summary}
\end{figure}

Clinical instruments in neurology are deployed in studies that span tens to hundreds of patients.
This reflects the realities of patient recruitment timescales, regulatory oversight, and the demands of longitudinal monitoring in complex disease populations.
Even the most coordinated multicentre efforts operate within these bounds: the Mobilise-D consortium, spanning six disease cohorts and sixteen clinical sites, enrolled on the order of 750 participants over several years.
The motivating question is therefore not how to mimic the scale of general-purpose wearables research, but how to learn robust, interpretable representations under strict clinical constraints.
While recent large-scale time-series foundation models (TSFMs)~\citep{Das2024TimesFM,Ansari2024Chronos,Liu2025TimerXL,Liu2024MoiraiMoE,Cohen2025Toto} achieve massive scale through out-of-domain consumer corpora or synthetic data mixtures, transferring these models to healthcare endpoints often compromises reliability.
Representations learned from domain-agnostic pretraining---whether on consumer activity corpora or synthetic simulations---need not encode the clinical structure on which interpretable, actionable inference depends; that structure must be built in from the outset through exposure to authentic patient variation.
In contrast, pretraining directly on in-domain clinical data---even when limited to hundreds of patients---forces the latent space to organize strictly around authentic physiological variation.
When the corpus is small, scale cannot rescue the model, but the resulting representations gain a critical property for clinical deployment: reliability and interpretability grounded in real patient biomechanics.

In clinical movement analysis, this translates into a concrete architectural discipline.
Every parameter must be allocated to learning the biomechanics underlying neurodegeneration---such as the shuffling hypokinesia of Parkinson's disease or the axial instability of cerebellar ataxia---rather than memorizing the MEMS noise floor, thermal bias drift, or the arbitrary yaw convention of a clip-on attachment.
Capacity in excess of the kinematic degrees of freedom of pathological gait is not merely wasted; it creates degrees of freedom in which sensor artefacts can stabilise as if they were clinically meaningful features.
A lightweight architecture is therefore a statistical requirement in this regime, not a deployment convenience.
One way to enforce it is to require the model to predict the future kinematic state of the body in a learned latent space, so that the representation is grounded in the dynamics of locomotion rather than the idiosyncrasies of the sensor.

This constraint also shapes the choice of sensing modality.
Clinical gait assessment with a single wearable requires a sensor that resolves the full kinematic state of trunk motion---both translational and rotational components.
A tri-axial accelerometer captures linear acceleration associated with foot-strike impact and gravitational loading, but is insensitive to the angular velocity that governs swing-phase progression, turning mechanics, and postural stabilisation.
These are not redundant signals: gait phase identification from accelerometry alone is often insufficient outside controlled laboratory settings~\citep{Lin2025GaitReview}, and many of the signatures most relevant to neurological assessment are rotational in nature.
Freezing of gait in Parkinson's disease is identified from a pathological power surge in the 3--8\,Hz band of lumbar acceleration~\citep{Moore2008FOG,Bachlin2010FOG}, and the loss of rhythmic lower-limb rotation that marks the episode is difficult to distinguish from a voluntary stop without angular velocity data.
Longitudinal kinematic features from combined inertial signals predict future fall risk in Parkinson's disease with AUC exceeding 0.90 at 24 months~\citep{Sotirakis2024FallsPD}, and stride-to-stride variability in gait dynamics is among the strongest predictors of functional decline in ageing populations~\citep{Hausdorff2007GaitDynamics}.
More broadly, the spastic gait of multiple sclerosis, the hypometric shuffling of Parkinson's disease, and the wide-based veering of cerebellar ataxia each produce distinct rotational trunk signatures that are difficult to differentiate without the gyroscope channel.
Six-axis measurement---tri-axial accelerometer fused with tri-axial gyroscope---follows from the biomechanics.

Sensor placement deserves the same deliberate treatment.
Most clinical protocols and field deployments are constrained to a single unit, so the question of where to place it carries real diagnostic weight.
The lower back, at or near the fifth lumbar vertebra, is the placement that returns the most clinical information from one sensor: it lies close to the body's centre of mass, and trunk acceleration during locomotion provides an accurate proxy for whole-body CoM displacement that is not recoverable from any other single site~\citep{Zijlstra2003CoM,MoeNilssen2004Gait}.
Chest-mounted sensors introduce respiratory contamination at approximately 0.3\,Hz with odd harmonics extending into 1.5--2.5\,Hz---precisely the bands that carry gait-cycle information---whereas lumbar signals reflect stride-driven CoM displacement with minimal respiratory confound~\citep{MoeNilssen1998,PatelPavic2019}.
The difference is clinically material: chest placement degrades the 3--8\,Hz freezing band on which closed-loop cueing systems depend, and cerebellar ataxia---fundamentally a disorder of axial coordination---propagates most cleanly through the lumbar sensor, where it is not obscured by upper-body compensatory motion.
Mobilise-D's adoption of lumbar placement across all six disease cohorts and sixteen clinical sites~\citep{Palmerini2023MobiliseD} reflects this field-wide consensus; \sonata's pretraining corpus follows the same standard.

Together, six-axis measurement and lumbar placement define three physics gates for corpus inclusion.
The first two are direct: the sensor must provide six-axis kinematics and must be trunk-mounted at the waist or lumbar spine.
The third is subtler: the data must preserve total rather than linear acceleration, because the gravitational component provides a fixed directional reference that reveals how the sensor is oriented relative to vertical---without it, translational acceleration and sensor tilt become indistinguishable.

These gates exclude the majority of available data: the largest open accelerometry benchmarks strip gravity by default; the richest wrist-mounted cohorts are off-body; prominent chest-IMU collections conflate respiratory and kinematic signals; gated institutional datasets remain inaccessible at time of writing.
What survives is nine open datasets spanning approximately 739 subjects and 190{,}000 windows~\citep{Anderson2026WearGait,Palmerini2023MobiliseD}: Parkinson's disease patients and age-matched controls, multiple sclerosis, stroke, and cerebellar and peripheral neuropathy cohorts, community-dwelling elderly fallers assessed prospectively, activities of daily living across healthy adult populations, and---critically---three sources with directional fall labels that capture the rare transient kinematics most consequential for fall-risk stratification.
The model makes no claim beyond this domain; what it has not been trained to learn, it should not be expected to represent.

\sonata is the pretraining framework that results from applying these constraints consistently.
It is a physics-informed world model for six-axis trunk kinematics, targeting the small-cohort clinical regime directly.
Where prior self-supervised methods rely on masked reconstruction or contrastive objectives over IMU sequences~\citep{Xu2021LIMUBERT,Haresamudram2021CPC,Xu2023UniHAR}, \sonata instead learns by predicting the future kinematic state of the body in a shared embedding space---following the latent-predictive paradigm recently established in biosignal and physical domains~\citep{Maes2026LeWM,Panchavati2026Laya}, and applying it here under the constraints of clinical data scale.
A model that can anticipate how a patient's trunk will move over the next half-second has, in some meaningful sense, internalised the coupled rotational-translational dynamics of human locomotion---gravity, stride cadence, turning dynamics, and the characteristic asymmetries that mark gait disorder---rather than the spectral fingerprint of any particular MEMS device---to the extent that sensor-specific noise is not predictive of future kinematic state.

This choice of objective aligns with a deeper property of the clinical problem itself.
Neurological disease progression is fundamentally dynamical: the relevant question is not whether a patient's gait is abnormal at a given instant, but whether their kinematic trajectory is diverging from their own baseline in a meaningful direction.
A world model is a natural fit for this problem in a way that purely reconstructive objectives may not be---a point we return to in the discussion.

The model comprises 3.77 million parameters in total, with hybrid backbone capacity matched to the kinematic degrees of freedom of pathological gait rather than to dataset volume.
This paper presents the \sonata pretraining corpus and methodology; downstream clinical transfer evaluation is the subject of a companion report.

\paragraph{Contributions.}
\begin{enumerate}[leftmargin=*, itemsep=2pt, parsep=0pt]
  \item \textbf{Architecture.} A 3.77\,M-parameter hybrid model interleaving global spectral filters with gated linear-recurrent state updates and cross-unit state mixing, purpose-built for six-axis trunk IMU kinematics under clinical data scarcity.
  \item \textbf{First application of latent world models to IMU kinematics.} Prior self-supervised IMU methods~\citep{Xu2021LIMUBERT,Haresamudram2021CPC,Xu2023UniHAR,Xu2025RelCon} rely on masked reconstruction or contrastive objectives that reward waveform fidelity rather than dynamical understanding. We adapt the Latent World Model objective~\citep{Maes2026LeWM} to inertial gait data and show, via controlled same-encoder ablations, that latent-state prediction with a linear head outperforms raw-signal forecasting on clinical discrimination and low-label transfer.
  \item \textbf{Evaluation protocol.} A 14-arm suite spanning clinical discriminability, noise robustness, cross-cohort transfer, and latent-structure diagnostics---designed to stress-test IMU representations beyond single-task accuracy.
  \item \textbf{Clinically motivated corpus curation.} Explicit inclusion criteria (six-axis, trunk-mounted, gravity-retained) defining a reproducible filtering methodology for clinical IMU pretraining across nine public sources (${\sim}$739 subjects, 190{,}000 windows).
\end{enumerate}

% sections/02_related_work.tex
% Related Work — tight, no subsections, bleeding-edge narrative

\section{Related Work}\label{sec:related_work}

Three strands of recent work frame the design space for \sonata:
wearable foundation models that demonstrate large-scale self-supervised
transfer across health endpoints; sub-quadratic hybrid architectures that
collapse the parameter cost of long-context modelling; and nascent world
models of the body that recast physiological forecasting as a generative
prior over kinematic and autonomic state.

\paragraph{Wearable foundation models.}
RelCon~\citep{Xu2025RelCon} trains on one billion accelerometry segments
from 87,376 participants using relative contrastive learning.
The Large Sensor Model~\citep{Narayanswamy2025LSM} scales this to 40 million
hours across six modalities and derives scaling laws for wearable sensing.
NormWear~\citep{Luo2025NormWear} handles arbitrary physiological signal
configurations through wavelet tokenisation and channel-aware attention across
18 downstream tasks.
SensorLM~\citep{Zhang2025SensorLM} bridges sensor data and natural language,
while AccelFM~\citep{Abbaspourazad2025AccelFM} distils PPG representations into
an accelerometer encoder predictive of 46 health targets.
Raw-IMU encoders represent a more proximate baseline: PRiMuS~\citep{Das2025PRiMuS}
leverages EgoExo4D with multimodal self-supervision; UniMTS~\citep{Zhang2024UniMTS}
aligns heterogeneous motion series with natural language; and
HiMAE~\citep{Lee2025HiMAE} applies hierarchical masked autoencoding---a
reconstruction objective that JEPA-style latent prediction directly supersedes.
Three structural limitations persist across all these works:
\emph{(i)} all rely on single-axis accelerometry, never the full 6-axis IMU
signal that captures rotational kinematics;
\emph{(ii)} pretraining corpora are predominantly proprietary, obstructing
reproducibility and clinical adaptation;
\emph{(iii)} populations with movement disorders---the primary clinical
beneficiaries of precise kinematic modelling---are absent from every
large-scale pretraining corpus.
\sonata{} does not compete with these models---it targets a regime they are
not designed for: six-axis trunk IMU in movement-disorder cohorts, where
the relevant clinical signal is rotational, data is limited to hundreds of
patients, and representations must be interpretable under strict regulatory
and deployment constraints.

\paragraph{World models for health signals.}
A recent capability survey~\citep{Qazi2025WMClinical} introduces a four-level
ladder for health world models (L1 temporal prediction to L4 closed-loop
planning) and identifies \emph{waveforms and multimodal sensor integration}
as open gaps at every level.
The Joint Embedding Predictive Architecture~\citep{Maes2026LeWM} addresses
this by predicting future \emph{latent} embeddings rather than reconstructing
the raw signal, discarding unpredictable noise while preserving semantic
structure.
This advantage has been validated across biosignals:
Laya~\citep{Panchavati2026Laya} applies LeJEPA to 29\,109 hours of EEG,
gaining $+$10.8\% seizure-detection accuracy and retaining 91\% clean
performance at 10\,dB SNR versus 68\% for reconstruction baselines;
EchoJEPA~\citep{Munim2026EchoJEPA} reduces LVEF estimation error by 26.7\%
over pixel-reconstruction for echocardiography;
medDreamer~\citep{Xu2025MedDreamer} adapts DreamerV3 to irregular EHRs,
reducing estimated sepsis mortality by 11\%;
and CLARITY~\citep{Ding2025CLARITY} shows latent-space dynamics outperform
diffusion-based generation by 9.2\% F1 for oncology treatment planning.
JETS~\citep{Xie2025JETS} applies JEPA to wearables but targets
daily-resolution aggregated metrics, not raw kinematic-rate inertial signals.
PULSE~\citep{Chen2025PULSE} frames physiological time series via dynamical
system representations but operates on non-IMU signals without action
semantics or JEPA-style latent prediction.
\sonata{} occupies the remaining vacancy: the first JEPA-based world model
for raw kinematic IMU.

\paragraph{Sub-quadratic sequence models and hybrid architectures.}
The delta rule, originally formulated as fast weight programming for linear attention~\citep{Schlag2021Linear}, provides a mechanism for targeted memory erasure.
DeltaNet~\citep{Yang2024DeltaNet} scales this principle via chunkwise parallelization, yielding a
Householder-structured state transition that selectively erases stale memory
at $O(Ld^2)$ cost.
Gated DeltaNet~\citep{Yang2025GatedDeltaNet} adds a Mamba-2-style~\citep{Dao2024Mamba2} scalar
decay gate, inheriting DeltaNet's targeted updates alongside Mamba-2's
efficient filtering on complementary synthetic tasks.
A systematic study of 72 hybrid models~\citep{Wang2025SystematicHybrid}
establishes that generation-3 delta-rule layers are the critical ingredient
for effective hybrids and that a 3:1 linear-to-full ratio maximises recall
at minimum compute cost---a finding validated at 48B parameters by Kimi
Linear~\citep{KimiTeam2025KimiLinear}, where the 3:1 ratio reduces KV cache
by 75\%, and grounded theoretically by OLMo Hybrid~\citep{Merrill2026OLMoHybrid}, which proves that such interleaved topologies achieve strict expressivity advantages (e.g., state-tracking in NC$^1$) over homogeneous stacks. Efficient distillation techniques (e.g., HALO~\citep{Chen2026HALO}) further streamline the construction of these massive hybrids, enabling extreme long-context scaling.
Recent extensions deepen expressivity: DeltaProduct~\citep{Siems2025DeltaProduct}
composes Householder reflections to access the full orthogonal group, and
EFLA~\citep{Lei2025EFLA} replaces Euler discretisation with an exact ODE
solution, reducing perplexity on standard language modelling benchmarks and
improving robustness to input noise.
On the time-series side, Reverso~\citep{Fu2026Reverso} interleaves long
convolution with DeltaNet blocks to produce 0.2M--2.6M-parameter TSFMs that
match Transformer-based models in zero-shot univariate forecasting.

\paragraph{The gap.}
No prior work applies hybrid delta-rule architectures to \emph{raw multivariate
inertial time series}, and no existing model adopts a JEPA-style latent
prediction objective for raw kinematic-rate IMU signals.
Reverso is the closest architectural analogue---same delta-rule backbone,
comparable scale---yet targets stationary univariate forecasting without
clinical populations.
Our first controlled baseline therefore keeps this hybrid backbone family fixed and asks a narrower question: in the low-data clinical regime, is it better to pretrain for latent-state prediction or for direct raw future-signal forecasting?
Laya is the nearest world-model predecessor for biosignals, but EEG carries
no action semantics; kinematic IMU admits gravity-referenced coordinate frames
and counterfactual intervention via altered motion parameters that EEG
structurally cannot support.
\sonata{} closes both gaps: a 3.77M-parameter hybrid model of spectral long
convolution and Gated DeltaNet, pretrained on a reproducible 6-axis corpus
spanning fall and neurodegenerative gait populations, with a JEPA pretraining
objective that enables action-conditioned kinematic rollouts for the first time.

% sections/04_data.tex
% Pretraining Corpus

\FloatBarrier
\section{Pretraining Corpus}\label{sec:data}

\begin{table*}[!t]
\centering
\caption{Pretraining corpus. Fall\,\% denotes the fraction of windows
  carrying a positive fall label for that source. For LTMM, fall-positive
  labels reflect retrospective faller-subject status rather than
  within-window impact events (see text).}
\label{tab:corpus_summary}
\small
\begin{tabular}{lrrrrl}
  \toprule
  Dataset & Subjects & Windows & Fall\,\% & Source\,Hz & Placement \\
  \midrule
  Voisard~\citep{Voisard2025GaitDataset}      & 260 & 15{,}467 &  0.0\% & 100     & Lower back \\
  WearGait-PD~\citep{Anderson2026WearGait}    & 184 & 16{,}739 &  0.0\% & 100     & Lower back \\
  Mobilise-D TVS~\citep{Palmerini2023MobiliseD}& 105 & 14{,}416 &  0.0\% & 100     & Lower back \\
  LTMM~\citep{Weiss2013LTMM}                 &  73 &  1{,}351 & 54.5\% & 100     & Lower back \\
  SisFall~\citep{Sucerquia2017SisFall}        &  38 & 84{,}634 & 75.9\% & 200     & Waist \\
  UCI HAR~\citep{Anguita2013UCIHAR}           &  30 &  5{,}125 &  0.0\% &  50     & Waist \\
  UMAFall~\citep{Casilari2017UMAFall}         &  19 &  9{,}161 & 73.2\% &  20     & Waist \\
  FallAllD~\citep{Saleh2021FallAllD}          &  15 & 32{,}641 & 68.2\% & 238     & Waist \\
  USC-HAD~\citep{Zhang2012USCHAD}             &  14 & 10{,}961 &  0.0\% & 100     & Waist \\
  \midrule
  \textbf{Total} & \textbf{$\sim$739} & \textbf{190{,}495} & \textbf{49.3\%} & & \\
  \bottomrule
\end{tabular}
\end{table*}

Pretraining a general-purpose IMU foundation model requires a corpus sufficiently diverse to
prevent spurious correlations arising from any single sensor configuration,
population, or activity distribution.
We curate a corpus of nine publicly available 6-axis IMU datasets
spanning clinical and laboratory settings across multiple countries
(Table~\ref{tab:corpus_summary}).
Three design constraints govern inclusion: (i)~full 6-axis sensing
(tri-axial accelerometer and tri-axial gyroscope), since the coupled
translational–rotational signal is necessary for learning the
gravity–motion decomposition; (ii)~waist or lower-back sensor placement,
ensuring a consistent biomechanical reference at the body's centre of mass;
and (iii)~total acceleration with gravity retained: beyond its role as an
orientation reference, the gravitational component encodes postural
information---trunk inclination and weight-bearing asymmetry---that is
directly predictive of fall risk and cannot be recovered once subtracted.
Gates (ii) and (iii) are complementary: gravity constrains pitch and roll
but leaves yaw undefined; the lumbar placement resolves this remaining
ambiguity through spinal anatomy, together fully constraining the sensor's
orientation in three dimensions.
Together, these constraints yield a corpus of ${\sim}739$ subjects and
190{,}495 five-second windows spanning activities of daily living, scripted
falls, and clinical gait assessments.
\inmode{Draft}{Full per-dataset provenance is given in Appendix~\ref{app:data_details}.}

\paragraph{Preprocessing.}
All sources are harmonised to a shared canonical form before training.
Each signal is first rotated into a common biomechanical reference frame
(ISB convention~\citep{Wu1995ISB}: vertical, anterior, medial axes), then resampled to
100\,Hz using polyphase anti-aliased filtering~\citep{Virtanen2020SciPy}.
The resampled signal is segmented into windows of 512 samples (5.12\,s),
each represented as a tensor of shape $(512, 6)$, using a 50\,\% stride
by default.
Channels are scaled to $[-1,\,1]$ using fixed constants matched to the
physical full-scale limits of common MEMS IMU hardware
($\pm16\,g$, $\pm2000\,^\circ\!/\!$s) rather than per-window statistics,
preserving the relative magnitude of the gravity vector as an always-available
postural cue.

\paragraph{Fall coverage.}
Scripted fall events, which are rare relative to activities of daily living,
are oversampled using a dense windowing pass with 90\,\% overlap over
fall-annotated regions---windows from this pass are labelled positive if
they intersect the annotated fall interval, making this an upsampling
heuristic rather than a guarantee that the impact event falls within each
window.
Falls are annotated with a four-class directional taxonomy (forward,
backward, lateral, near-fall) and a binary \texttt{fall\_mask} field that
aggregates all four classes for use by class-balanced samplers in
fine-tuning.
LTMM does not contain scripted fall events; its windows are labelled by
prospective faller-subject status and contribute supervision signal for
fall-risk stratification rather than fall-event detection.

\paragraph{Data augmentation.}
The nominal production configuration applies four on-the-fly
transformations~\citep{Um2017Data,Iwana2021Empirical} chosen to reflect the physical variability of real-world
sensor deployments: sagittal parity reflection, SO(3) placement tilt
($\sigma\!=\!8^\circ$, simulating clip-on re-placement error), gyro bias
drift (Brownian motion, modelling MEMS gyroscope instability), and
percentile censoring.
\inmode{Draft}{Augmentation parameters and ablation results are reported in
Appendix~\ref{app:augmentation}.}

% sections/03_architecture.tex
% Architecture — Sonata hybrid LongConv / Gated DeltaNet

\section{Architecture and Objective Function}\label{sec:architecture}

\begin{figure*}[!t]
\centering
\begin{tikzpicture}[
  %% ---- Tensor-stack base style (tall thin block = sequence tensor) ----
  tstack/.style={
    minimum width=0.46cm, minimum height=1.62cm,
    rounded corners=1.5pt, inner sep=0pt
  },
  %% ---- Backbone block styles (tall, matching tensor-stack height) ----
  blockC/.style={
    draw=ttorange!80!black, fill=ttorange!28,
    minimum width=0.50cm, minimum height=1.62cm,
    rounded corners=2pt, font=\scriptsize\bfseries, inner sep=1pt
  },
  blockG/.style={
    draw=blue!65, fill=blue!22,
    minimum width=0.50cm, minimum height=1.62cm,
    rounded corners=2pt, font=\scriptsize\bfseries, inner sep=1pt
  },
  %% ---- Pipeline stage boxes (same height as blocks) ----
  pipe/.style={
    draw=gray!52, fill=gray!10, rounded corners=3pt,
    minimum height=1.62cm, align=center, font=\scriptsize, inner sep=4pt
  },
  pipeSonatablue/.style={
    draw=sonatablue!80!black, fill=sonatablue!16, rounded corners=3pt,
    minimum height=1.62cm, align=center, font=\scriptsize, inner sep=4pt
  },
  %% ---- Detail-panel operation boxes ----
  op/.style={
    draw=gray!52, fill=gray!10, rounded corners=2pt,
    minimum height=0.54cm, align=center, font=\tiny, inner sep=3pt
  },
  opO/.style={
    draw=ttorange!65, fill=ttorange!16, rounded corners=2pt,
    minimum height=0.54cm, align=center, font=\tiny, inner sep=3pt
  },
  opB/.style={
    draw=blue!55, fill=blue!12, rounded corners=2pt,
    minimum height=0.54cm, align=center, font=\tiny, inner sep=3pt
  },
  %% ---- Arrows ----
  arr/.style={-{Latex[length=3.5pt,width=2.5pt]}, thick},
  sarr/.style={-{Latex[length=2.5pt,width=2pt]}, semithick},
  darr/.style={-{Latex[length=3.5pt,width=2.5pt]}, thick, dashed, gray!55},
]

%% ================================================================
%% PANEL A --- Full forward-pass pipeline (Laya-style tall blocks)
%% ================================================================
\begin{scope}[yshift=5.80cm]

  %% ---- Input ----
  \node[pipe, minimum width=0.90cm] (X)
    {$\mathbf{X}$\\\tiny$B{\times}L{\times}6$};

  %% ---- Backbone makeup annotation ----
  \node[font=\scriptsize\itshape, gray!70, anchor=south]
    at (4.2, 1.16) {Unit backbone: 12x (SeqBlock + MLP)};

  %% ---- Norm + Embed ----
  \node[pipeSonatablue, minimum width=1.52cm, right=0.32cm of X] (nemb)
    {Norm\,\&\\Embed};

  %% ---- Embedded-sequence tensor stack (sonatablue) ----
  \node[tstack, draw=sonatablue!80!black, fill=sonatablue!28, right=0.28cm of nemb,
        label={[font=\tiny, gray!55]below:$L{\times}d$}] (ts0) {};
  \hlines{ts0}

  %% ---- Unit 1: [C, C, C, G] ----
  \node[blockC, right=0.36cm of ts0] (c0) {C};
  \node[blockC, right=0.13cm of c0]  (c1) {C};
  \node[blockC, right=0.13cm of c1]  (c2) {C};
  \node[blockG, right=0.13cm of c2]  (g3) {G};

  %% ---- Tensor stack after Unit 1 (blue = processed representation) ----
  \node[tstack, draw=blue!55, fill=blue!14, right=0.34cm of g3,
        label={[font=\tiny, gray!55]below:$\mathbf{H}^{(4)}$}] (ts1) {};
  \hlines{ts1}

  %% ---- Unit 2: [C, C, C, G] ----
  \node[blockC, right=0.34cm of ts1] (c4) {C};
  \node[blockC, right=0.13cm of c4]  (c5) {C};
  \node[blockC, right=0.13cm of c5]  (c6) {C};
  \node[blockG, right=0.13cm of c6]  (g7) {G};

  %% ---- Tensor stack after Unit 2 ----
  \node[tstack, draw=blue!55, fill=blue!14, right=0.34cm of g7,
        label={[font=\tiny, gray!55]below:$\mathbf{H}^{(8)}$}] (ts2) {};
  \hlines{ts2}

  %% ---- Unit 3: [C, C, C, G] ----
  \node[blockC, right=0.34cm of ts2] (c8)  {C};
  \node[blockC, right=0.13cm of c8]  (c9)  {C};
  \node[blockC, right=0.13cm of c9]  (c10) {C};
  \node[blockG, right=0.13cm of c10] (g11) {G};

  %% ---- Latent readout S_T (compact deep-blue stack) ----
  \node[tstack, draw=blue!65, fill=blue!25, minimum width=0.52cm,
        right=0.40cm of g11,
        label={[font=\tiny, gray!55]below:$\mathbf{S}_t$}] (zr) {};
  \hlines{zr}

  %% ---- JEPA Pre-training Branch (Top) ----
  \node[pipe, fill=green!12, draw=green!55, minimum height=0.64cm, minimum width=1.34cm, above=1.05cm of zr, anchor=south] (proj)
    {Projector $\pi_\phi$};
  \node[pipe, fill=green!12, draw=green!55, minimum height=0.64cm, minimum width=1.34cm, right=0.32cm of proj] (pred)
    {Predictor $W$};
  \node[font=\scriptsize, right=0.28cm of pred] (zh)
    {$\hat{\mathbf{S}}_{t+1}$};

  \draw[arr, dashed, gray!60] (zr.north) -- ++(0,0.60) -| (proj.south);
  \draw[arr, dashed, gray!60] (proj.east) -- (pred.west);
  \draw[arr, dashed, gray!60] (pred.east) -- (zh.west);

  \node[font=\tiny\itshape, gray!50, above=0.05cm of proj] {Pre-training};

  %% ---- Downstream linear probe ----
  \node[pipe, minimum width=1.34cm, right=0.32cm of zr] (dhead)
    {Linear\\probe};

  %% ---- Output ----
  \node[pipe, minimum width=0.90cm, right=0.28cm of dhead] (Yh)
    {$\hat{\mathbf{y}}$\\\tiny task pred.};

  %% ---- Forward arrows ----
  \foreach \a/\b in {
      X/nemb, nemb/ts0, ts0/c0,
      c0/c1, c1/c2, c2/g3, g3/ts1, ts1/c4,
      c4/c5, c5/c6, c6/g7, g7/ts2, ts2/c8,
      c8/c9, c9/c10, c10/g11,
      g11/zr, zr/dhead, dhead/Yh}
    \draw[arr] (\a) -- (\b);

  %% ---- State-weave arcs (bottom route, blue) ----
  \draw[arr, blue!60, rounded corners=3pt]
    (g3.south) -- ++(0,-0.58) -| (c4.south)
    node[near end, below, font=\tiny, blue!65] {state weave};
  \draw[arr, blue!60, rounded corners=3pt]
    (g7.south) -- ++(0,-0.58) -| (c8.south)
    node[near end, below, font=\tiny, blue!65] {state weave};

  %% ---- Unit grouping boxes (dashed borders, drawn after nodes) ----
  \draw[rounded corners=3pt, draw=gray!38, dashed]
    ($(c0.north west)+(-0.18,0.24)$) rectangle ($(g3.south east)+(0.18,-0.24)$);
  \node[font=\tiny, gray!55]
    at ($(c0.north)!0.5!(g3.north)+(0,0.40)$) {Unit~1};

  \draw[rounded corners=3pt, draw=gray!38, dashed]
    ($(c4.north west)+(-0.18,0.24)$) rectangle ($(g7.south east)+(0.18,-0.24)$);
  \node[font=\tiny, gray!55]
    at ($(c4.north)!0.5!(g7.north)+(0,0.40)$) {Unit~2};

  \draw[rounded corners=3pt, draw=gray!38, dashed]
    ($(c8.north west)+(-0.18,0.24)$) rectangle ($(g11.south east)+(0.18,-0.24)$);
  \node[font=\tiny, gray!55]
    at ($(c8.north)!0.5!(g11.north)+(0,0.40)$) {Unit~3};

\end{scope} %% end Panel A

%% Divider (removed)

%% ================================================================
%% PANEL B --- LongConvBlock detail
%% Width budget: box [-0.12, 8.22] = 8.34 cm
%% y_gate = 2.74,  y_main = 2.00
%% ================================================================
\begin{scope}

  %% Panel box (drawn first so nodes sit on top)
  \draw[rounded corners=4pt, draw=ttorange!42, fill=ttorange!5]
    (-0.12, 1.24) rectangle (7.90, 3.50);

  \node[font=\small\bfseries, ttorange!80!black, anchor=west]
    at (0.02, 3.68) {LongConvBlock};

  %% Input
  \node[op, minimum width=0.52cm] (lci) at (0.38, 2.00) {$\mathbf{x}$};

  %% Gate branch
  \node[opO, minimum width=1.32cm] (gate) at (2.42, 2.74) {Gate\;$\mathbf{g}$};

  %% Gate multiply (main path)
  \node[op, minimum width=0.44cm] (gm) at (2.42, 2.00) {${\odot}$};

  %% Main-path operations
  \node[opO, minimum width=1.08cm] (fft) at (3.90, 2.00) {$\mathbf{w}{\ast}(\cdot)$};
  \node[op,  minimum width=0.64cm] (rl)  at (5.18, 2.00) {ReLU};
  \node[op,  minimum width=0.44cm] (ad)  at (6.14, 2.00) {$+$};
  \node[op,  minimum width=0.44cm] (ln)  at (6.92, 2.00) {LN};
  \node[font=\tiny, anchor=west]   (lco) at (7.48, 2.00) {out};

  %% Arrows --- gate branch
  \draw[sarr] (lci.north) |- (gate.west);
  \draw[sarr] (gate.south) -- (gm.north);

  %% Arrows --- main path
  \draw[sarr] (lci.east) -- (gm.west);
  \draw[sarr] (gm)  -- (fft);
  \draw[sarr] (fft) -- (rl);
  \draw[sarr] (rl)  -- (ad);
  \draw[sarr] (ad)  -- (ln);
  \draw[sarr] (ln)  -- (lco);

  %% Residual
  \draw[sarr, gray!45] (lci.south) -- ++(0,-0.34) -| (ad.south);

  %% rFFT annotation
  \node[font=\scriptsize\itshape, gray!50, anchor=north] at (3.90, 1.68)
    {rFFT conv};

\end{scope} %% end Panel B

%% ================================================================
%% PANEL C --- GatedDeltaNetBlock detail
%% xshift = 8.16 cm
%% Width budget: box [-0.12, 9.46] = 9.58 cm
%% Global right edge: 8.16 + 9.46 = 17.62 cm
%% y_gate = 2.96,  y_main = 1.96
%%
%% Layout rationale: give the state-update box more width and split the
%% recurrence across three lines so the panel reads less squashed at print
%% size. The SiLU gate still sits directly above the multiply node (od2).
%% ================================================================
\begin{scope}[xshift=8.16cm]

  %% Panel box (drawn first)
  \draw[rounded corners=4pt, draw=blue!38, fill=blue!4]
    (-0.12, 1.16) rectangle (9.72, 3.60);

  \node[font=\small\bfseries, blue!70!black, anchor=west]
    at (0.02, 3.78) {GatedDeltaNetBlock};

  %% Input
  \node[op, minimum width=0.52cm] (gdi) at (0.42, 1.96) {$\mathbf{x}_t$};

  %% Q/K/V/beta projections
  \node[opB, minimum width=1.08cm] (prj) at (1.70, 1.96)
    {$\mathbf{q},\mathbf{k},\mathbf{v},\beta$};

  %% Decay gate (branch above state recurrence)
  \node[opB, minimum width=1.32cm] (gt) at (3.96, 2.96) {Decay\;$g_t$};

  %% State recurrence
  \node[opB, minimum width=2.62cm] (src) at (3.96, 1.96)
    {$\mathbf{S}_t = g_t\!\cdot\!\mathbf{S}_{t-1}\bigl($\\
     $\mathbf{I}-2\beta_t\mathbf{k}_t\mathbf{k}_t^\top\bigr)$\\
     $+\beta_t\mathbf{v}_t\mathbf{k}_t^\top$};

  %% Readout: keep a distinct slot between recurrence and output gate
  \node[opB, minimum width=1.02cm] (rmn) at (6.36, 1.96)
    {RMSNorm\\$(\mathbf{S}_t\mathbf{q}_t)$};

  %% Output gate (multiply)
  \node[op, minimum width=0.38cm] (od2) at (7.62, 1.96) {$\odot$};

  %% SiLU output gate sits directly above the multiply node
  \node[op, minimum width=1.18cm] (slg) at (7.62, 2.96)
    {SiLU$(W_z\mathbf{x}_t)$};

  %% Add, LN, out
  \node[op, minimum width=0.38cm] (ad2) at (8.42, 1.96) {$+$};
  \node[op, minimum width=0.38cm] (ln2) at (9.10, 1.96) {LN};
  \node[font=\tiny, anchor=west]  (gdo) at (9.40, 1.96) {out};

  %% Arrows
  \draw[sarr] (gdi) -- (prj);
  \draw[sarr] (gdi.north) |- (gt.west);
  \draw[sarr] (gt.south) -- (src.north);
  \draw[sarr] (prj) -- (src);
  \draw[sarr] (src) -- (rmn);
  \draw[sarr] (rmn) -- (od2);
  \draw[sarr] (slg.south) -- (od2.north);
  \draw[sarr] (od2) -- (ad2);
  \draw[sarr] (ad2) -- (ln2);
  \draw[sarr] (ln2) -- (gdo);

  %% Residual
  \draw[sarr, gray!45] (gdi.south) -- ++(0,-0.38) -| (ad2.south);

\end{scope} %% end Panel C

\end{tikzpicture}
\caption{
  \textbf{\sonata architecture, pretraining objective, and downstream probe.}
  \emph{Top:} The input sequence is embedded into $\mathbf{H}^{(0)}$ and processed by a hybrid backbone interleaving LongConvBlocks (\textbf{C}) and GatedDeltaNetBlocks (\textbf{G}).
  Intermediate recurrent states weave across units via state injection (\cref{eq:state_weave}).
  During self-supervised pretraining, the terminal latent state $\mathbf{S}_t$ is routed through an expand--compress projector $\pi_\phi$ and linear predictor $W_p$ to minimize the Latent World Model (LWM) objective (\cref{alg:lwm_pretraining}).
  For downstream tasks, the LWM heads are discarded, the encoder is frozen, and the final MLP-block embeddings are mean-pooled over time before a lightweight logistic-regression probe is applied.
  \emph{Bottom:} Detailed schematics of the LongConvBlock \citep{Fu2026Reverso} and GatedDeltaNetBlock \citep{Yang2025GatedDeltaNet} comprising the hybrid sequence mixer. Full operational details and mathematical formulations are provided in \cref{subsec:backbone}.
}
\label{fig:arch}
\end{figure*}
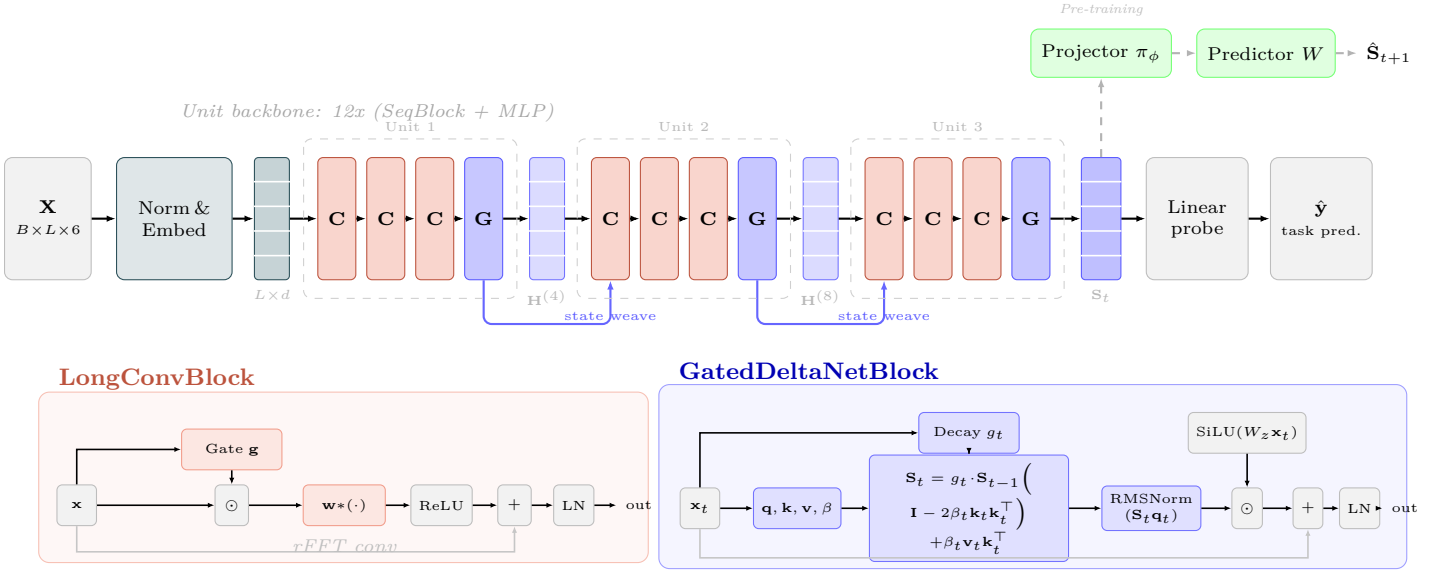

% Full architectural figure  ---  Laya-inspired visual style
% -----------------------------------------------------------------------
%
% Width budget (textwidth ~18 cm):
%   Panel A pipeline: ~17.1 cm (relative-positioned nodes)
%   Panel B (LongConv): box [-0.12, 8.22] = 8.34 cm  (xshift 0)
%   Panel C (GDN):      xshift 8.60, box [-0.12, 8.88] = 9.00 cm
%   Total bottom: 8.60 + 9.00 = 17.60 cm
%

% -----------------------------------------------------------------------

% -----------------------------------------------------------------------
\subsection{A parameter-efficient world model for human motion}%
\label{subsec:arch_overview}
% -----------------------------------------------------------------------

\sonata is a hybrid time-series foundation model designed for six-axis waist and lower-back IMU kinematics at 100\,Hz, interleaving long convolutions with Gated DeltaNet blocks. 
Building upon the hybrid architecture of \citet{Fu2026Reverso}, this formulation achieves Pareto-optimal representation quality, matching the performance of substantially larger architectures at a fraction of the parameter cost.
The efficacy of this hybrid paradigm is corroborated by recent large-scale evaluations in language modeling \citep{KimiTeam2025KimiLinear}, demonstrating that selective recurrence combined with convolutional context aggregation efficiently captures both local associations and global dependencies.

Following the conceptual framework of \citet{Maes2026LeWM}, \sonata is cast as an \emph{actionable world model} that compresses the underlying physics of human locomotion into a compact, predictable latent space.
The architectural capacity is strictly allocated toward learning biomechanical structures indicative of neurodegeneration---such as the shuffling hypokinesia of Parkinson's disease or the wide-based axial instability of cerebellar ataxia---rather than reconstructing high-variance sensor artefacts.
This objective motivates three primary design choices: (i) the retention of six-axis inertial inputs including gravity (\cref{sec:data}) to enable the disentanglement of postural tilt from dynamic acceleration; (ii) a latent-predictive self-supervised training objective; and (iii) a highly parameter-efficient total parameter budget (${\approx}$3.77\,M) detailed in \cref{tab:config}.
Specifically, sensor data are normalized and embedded into $\mathbb{R}^{L\times d}$ before passing through a 12-layer $[\text{C,C,C,G}]\times 3$ hybrid backbone---where \textbf{C} denotes a Long-convolution block and \textbf{G} denotes a Gated DeltaNet block---terminating in a latent read-out that feeds into lightweight pretraining heads and a frozen linear probe for downstream evaluation (\cref{fig:arch}). 
The primary mathematical notation utilized throughout this architecture is summarized in \cref{tab:symbols}.

% -----------------------------------------------------------------------
\subsection{Hybrid backbone: long convolutions for spectra, gated delta
  for state}\label{subsec:backbone}

% -----------------------------------------------------------------------

\paragraph{Long-convolution block (C).}
Each C block implements a gated long convolution following
Reverso~\citep{Fu2026Reverso}: a lightweight data-dependent gate
(comprising depthwise convolution, SiLU activation, pointwise convolution, and sigmoid) selectively suppresses frequency bands before filtering,
modulating the input to ignore high-frequency noise or electromyography (EMG) artefacts on a per-input basis.
The modulated signal is then convolved via rFFT circular convolution
with a per-channel learnable kernel $\mathbf{w}\in\mathbb{R}^{d\times L}$,
followed by a residual LayerNorm.
Complexity is $O(dL\log L)$ per block.
In the context of IMU gait, a stride cycle at 1--2\,Hz constitutes a global periodic pattern spanning the full 5.12\,s window. 
A full-length learnable filter naturally represents the stride fundamental and its harmonics as a single spectral template---an operation impossible for local convolutions with restrictive receptive fields.
The three stacked C blocks per sequence unit build a deep filter bank. 
Each layer applies a distinct spectral filter, enabling the network to jointly represent multi-scale temporal patterns (e.g., stride cycle, within-stride phase, step asymmetry) without compressing or discarding positional information.
Essentially, the convolutional layers collectively answer: \emph{"What frequencies and patterns exist within this window?"}

\paragraph{Gated DeltaNet block (G).}
The recurrent blocks use multi-head Gated
DeltaNet~\citep{Yang2025GatedDeltaNet}. Formally, given an input token $\mathbf{x}_t \in \mathbb{R}^d$, the block first computes linear projections for the query, normalized key, value, and update rate:
\begin{equation}\label{eq:gdn_projections}
  \mathbf{q}_t = W_q \mathbf{x}_t, \quad
  \mathbf{k}_t = \frac{W_k \mathbf{x}_t}{\|W_k \mathbf{x}_t\|_2}, \quad
  \mathbf{v}_t = W_v \mathbf{x}_t, \quad
  \beta_t = \sigma(W_\beta \mathbf{x}_t).
\end{equation}
The associative state $\mathbf{S}_t\in\mathbb{R}^{d_v/H \times d_k}$ then evolves via the gated delta rule~\citep{Yang2024DeltaNet}:
\begin{equation}\label{eq:gdn_recurrence}
  \mathbf{S}_t
    = g_t\!\cdot\!\mathbf{S}_{t-1}\!\bigl(
        \mathbf{I} - 2\beta_t\mathbf{k}_t\mathbf{k}_t^\top
      \bigr)
      + \beta_t\mathbf{v}_t\mathbf{k}_t^\top.
\end{equation}
The Householder factor $\mathbf{I}-2\beta_t\mathbf{k}_t\mathbf{k}_t^\top$
erases the component of $\mathbf{S}_{t-1}$ aligned with the incoming key
before writing the new association, enabling targeted forgetting unavailable
in additive linear attention.
A Mamba-2-style~\citep{Dao2024Mamba2} content-dependent decay gate $g_t = \sigma(W_g \mathbf{x}_t) \in (0,1)$ provides a
complementary global reset: at gait-phase boundaries, freezing-of-gait
episodes, and session transitions---the contexts where stale kinematic
state is most misleading---the gate is positioned to drive memory
clearance, enabling the model to adapt to clinical non-stationarity
without per-subject finetuning~\citep{Yang2025GatedDeltaNet}.

A more subtle design choice concerns the eigenspectrum of the
state-transition operator.
Classical DeltaNet confines it to $[0,1]$, which is adequate for monotone
dynamics but cannot represent sign-alternating behaviour.
\citet{Grazzi2025StateTracking} show that lifting the eigenspectrum to
$[-1,1]$ turns the linear recurrence into an expressive \emph{state-tracker};
for IMU kinematics this is not cosmetic---gait is a multi-scale periodic
process, from bilateral stride alternation to tremor and turning dynamics,
whose faithful latent representation requires sign-alternating recurrent
modes~\citep{Grazzi2025StateTracking}; restricting the eigenspectrum to
non-negative values measurably degrades clinical discrimination, as we
confirm empirically in \cref{subsec:eigval_ablation}.
\sonata enables this extended eigenspectrum in all G blocks.
The block reads out via a SiLU-gated FusedRMSNorm projection before a
residual LayerNorm.

While the convolutional layers comprehensively process the full sequence, they treat each window independently.
In contrast, the GDN accumulates a compressed kinematic summary over the 512 positions. 
At each intermediate G layer (positions 3 and~7 in the stack), the
terminal hidden state is injected into position~0 of the following layer:
\begin{equation}\label{eq:state_weave}
  \mathbf{H}^{(j+1)}_{\cdot,0,\cdot}
    \leftarrow
  \mathbf{H}^{(j+1)}_{\cdot,0,\cdot}
  + \mathbf{H}^{(j+1)}_{\cdot,L-1,\cdot},\quad j\in\{3,7\},
\end{equation}
providing lightweight recurrent coupling across backbone units without explicit state caching.
This explicitly supplies the subsequent convolutional unit with a compressed history of the preceding one.
Building upon the convolutional layers' spectral analysis, the GDN essentially answers: \emph{"Given everything that happened in this window, what is the current overarching kinematic state of the body?"}

\paragraph{Hybrid sequence modelling.}
The layer sequence $[\text{C,C,C,G}]\times 3$ adapts the architecture of \citet{Fu2026Reverso}, aggregating three global long-convolution layers upstream of each recurrent block. 
This conv-heavy configuration ensures that the Gated DeltaNet modules process spectrally organised representations rather than raw embeddings, so that recurrent capacity is directed toward temporal compression and state management rather than spectral feature extraction.
Additionally, each sequence block is coupled with a residual gated-MLP channel mixer. 
As the convolutional and recurrent blocks operate exclusively on a per-channel or per-head basis, the MLP serves as the sole mechanism for cross-channel feature combination (e.g., fusing vertical acceleration with pitch velocity).
Finally, input channels are normalised to $[-1,1]$ using fixed
physical-range constants (\cref{sec:data}), preserving the relative
magnitude of the gravity vector as an always-available postural
reference for resolving orientation in clinical populations.

% -----------------------------------------------------------------------
\subsection{Latent read-out and heads}\label{subsec:decoder}

\begin{algorithm}[!t]
\caption{Latent World Model (LWM) Pretraining Step}
\label{alg:lwm_pretraining}
\begin{algorithmic}[1]
\Require Context window $\mathbf{X}_t$, Target window $\mathbf{X}_{t+1}$
\Require Encoder $f_\theta$, Projector $\pi_\phi$, Predictor $W_p$
\State \textbf{Forward Pass:}
\State $\mathbf{H}_t \gets f_\theta(\mathbf{X}_t)$ \Comment{Encode context window}
\State $\mathbf{H}_{t+1} \gets f_\theta(\mathbf{X}_{t+1})$ \Comment{Encode target window (shared weights)}
\State $\mathbf{S}_t \gets (\mathbf{H}_t)_{\cdot,L-1,\cdot}$ \Comment{Context latent readout}
\State $\mathbf{S}_{t+1} \gets (\mathbf{H}_{t+1})_{\cdot,L-1,\cdot}$ \Comment{Target latent readout}
\State $\mathbf{Z}_t \gets \pi_\phi(\mathbf{S}_t)$ \Comment{Project context}
\State $\mathbf{Z}_{t+1} \gets \pi_\phi(\mathbf{S}_{t+1})$ \Comment{Project target}
\State $\hat{\mathbf{S}}_{t+1} \gets W_p\,\mathbf{Z}_t$ \Comment{Predict target latent state}
\State
\State \textbf{Objective Computation:}
\State $\mathcal{L}_\text{pred} \gets \| \hat{\mathbf{S}}_{t+1} - \mathbf{Z}_{t+1} \|_2^2$ \Comment{Prediction MSE loss}
\State $\mathcal{L}_\text{SIGReg} \gets \text{Epps--Pulley}\bigl([\mathbf{Z}_t;\,\mathbf{Z}_{t+1}],\,\mathcal{N}(0, I)\bigr)$ \Comment{Isotropic Gaussian penalty}
\State $\mathcal{L}_\text{LWM} \gets (1-\lambda)\mathcal{L}_\text{pred} + \lambda\mathcal{L}_\text{SIGReg}$ \Comment{Total objective}
\State
\State \textbf{Gradient Flow ($\leftarrow$):}
\State $\nabla \theta, \nabla \phi, \nabla W_p \gets \text{Backprop}(\mathcal{L}_\text{LWM})$ \Comment{End-to-end backpropagation}
\State \Comment{Note: No stop-gradients or EMA updates are used.}
\State $\theta \gets \theta - \eta \nabla \theta$ \Comment{Update Encoder}
\State $\phi \gets \phi - \eta \nabla \phi$ \Comment{Update Projector}
\State $W_p \gets W_p - \eta \nabla W_p$ \Comment{Update Predictor}
\end{algorithmic}
\end{algorithm}

% -----------------------------------------------------------------------

\paragraph{Latent read-out.}
During JEPA pretraining, the primary latent state is the terminal
sequence position of the final recurrent block:
\begin{equation}\label{eq:latent_readout}
  \mathbf{S}_T = \mathbf{H}^{(12)}_{\cdot,\,L-1,\,\cdot} \in \mathbb{R}^{B \times d},
\end{equation}
which serves as the projected prediction target for the latent world-model objective.
For the frozen-encoder downstream evaluation, however, embeddings are extracted one sublayer deeper from the final MLP block and mean-pooled over time:
\begin{equation}\label{eq:downstream_embedding}
  \mathbf{e}
    = \frac{1}{L}\sum_{t=1}^{L}\mathbf{H}^{(\mathrm{mlp},12)}_{\cdot,\,t,\,\cdot}
    \in \mathbb{R}^{B \times d},
\end{equation}
yielding the $(B,128)$ representation used by all downstream linear probes for every arm.

\paragraph{Latent World Model (LWM) objective.}
We optimize \sonata via a Latent World Model (LWM) loss, following the self-supervised frameworks established by \citet{Balestriero2025LeJEPA} and \citet{Maes2026LeWM}.
In domains characterised by high-variance physiological and sensor artefacts, signal-space objectives that reward direct waveform fidelity can force the encoder to memorize noise rather than extracting the underlying clinical state \citep{Panchavati2026Laya}.
More fundamentally, the choice of supervisory target creates an implicit information bottleneck: an encoder optimized to predict the next raw waveform is rewarded for high-frequency fidelity---reproducing transient sensor jitter, EMG crosstalk, and MEMS artefacts---whereas an encoder optimized to predict the next latent state is rewarded for cross-window predictability rather than intra-window fidelity---a pressure that favours structure persisting across consecutive windows, such as stride timing, postural asymmetry, and disease-specific gait signatures, over transient artefacts that do not.
The empirical consequence is that forecasting accuracy and clinical discrimination are not merely different metrics but competing demands on encoder capacity (\cref{tab:main_results}).
To circumvent this, the LWM objective eschews signal-level reconstruction in favor of predicting the latent state of a future context window directly from the current latent state.

Historically, Joint-Embedding Predictive Architectures (JEPAs) prevented representation collapse through an unprincipled patchwork of architectural heuristics, notably asymmetric exponential moving average (EMA) target encoders and stop-gradients; omitting these typically induced immediate collapse.

However, \citet{Balestriero2025LeJEPA} recently proved that the isotropic Gaussian is the uniquely optimal embedding distribution to minimise downstream prediction risk---specifically, bias and variance for linear probes.
Consequently, \sonata adopts their bespoke, heuristic-free formulation: it operates entirely end-to-end without stop-gradients or EMA targets.
Given consecutive context and target windows $(\mathbf{X}_t, \mathbf{X}_{t+1})$, the shared encoder $f_\theta$ produces latent states $\mathbf{S}_t$ and $\mathbf{S}_{t+1}$. The overarching objective collapses to a single two-term loss:
\begin{equation}\label{eq:lwm_loss}
  \mathcal{L}_\text{LWM} = (1-\lambda)\,\mathcal{L}_\text{pred} + \lambda\,\mathcal{L}_\text{SIGReg}.
\end{equation}
Here, the prediction loss is the mean squared error in the projected space:
\begin{equation}
  \mathcal{L}_\text{pred} = \| \hat{\mathbf{S}}_{t+1} - \pi_\phi(\mathbf{S}_{t+1}) \|_2^2,
\end{equation}
where $\hat{\mathbf{S}}_{t+1} = W_p\,\pi_\phi(\mathbf{S}_t)$ is the predictor's output.
$\mathcal{L}_\text{SIGReg}$ is the Epps--Pulley-based isotropic Gaussian
regulariser of \citet{Balestriero2025LeJEPA}; we refer the reader to
their formulation and provide empirical sensitivity analysis in
\cref{subsec:sigreg_ablation}.
Representation collapse is provably averted through the Sketched Isotropic Gaussian Regularization penalty ($\mathcal{L}_\text{SIGReg}$). 
Relying on the Cramér--Wold theorem---which states that a multidimensional distribution is uniquely determined by its one-dimensional marginals---SIGReg evaluates the empirical characteristic function of the projected batch embeddings along $M$ random one-dimensional projections.
By minimizing the Epps--Pulley normality test statistic between these projections and a standard Gaussian, SIGReg explicitly and stably guides the embedding space toward the optimal isotropic Gaussian distribution.
Crucially, the Epps--Pulley statistic is utilized because it possesses uniformly bounded gradients and curvature, preventing the instabilities inherent to moment-based or CDF-based normality tests.
The gradient flow throughout this architecture during pretraining is entirely symmetric and summarized in \cref{alg:lwm_pretraining}.

\paragraph{Controlled objective comparison.}
To isolate the role of the pretraining objective rather than the encoder family, we compare the production TTL-TSFM \textsc{LIN} variant against a causal raw-signal forecasting baseline denoted \textsc{MAE}, where the acronym refers to the mean-absolute-error training loss rather than to masked autoencoding. The shared component is the same hybrid LongConv + Gated DeltaNet encoder backbone, with the same frozen linear-probe downstream protocol; the only architectural divergence is in the pretraining head and supervisory signal. In \textsc{LIN}, two consecutive 512-step windows are encoded into terminal latent states, passed through the shared projector $\pi_\phi$, and coupled by a single linear predictor $W_p$ under the two-term latent loss $\mathcal{L}_\text{LWM}$. In \textsc{MAE}, the first 464 samples of a window are presented as context, zero-padded to length 512 for the FFT-based backbone, and the held-out final 48 samples are predicted directly in raw signal space by the standard \texttt{DecoderHead}: 48 learned query positions cross-attend over the encoder sequence states and are optimized with an L1 loss on the future 6-axis waveform. For downstream evaluation, all five arms are instead compared using the shared mean-pooled final-MLP embedding in \cref{eq:downstream_embedding}, rather than the terminal recurrent readout $\mathbf{S}_T$ from \cref{eq:latent_readout}. The point of the comparison is therefore not to introduce a second architectural story, but to test whether latent world-model pretraining yields more useful frozen representations in the low-data clinical regime than direct raw-signal forecasting of the broader time-series family exemplified by Reverso~\citep{Fu2026Reverso}. Our working hypothesis is that, under data scarcity, efficient pretraining should spend capacity on the biomechanics of disease progression rather than on the raw sensor noise floor or transient jitter.

\paragraph{Pretraining heads.}
To compute the prediction loss and apply the SIGReg regularizer, the latent state $\mathbf{S}_t$ is routed through an expand--compress MLP projector $\pi_\phi$.
Introduced theoretically by \citet{Balestriero2025LeJEPA} as a critical component for stabilizing self-supervised embeddings, this projector expands the latent state into a higher-width hidden layer before returning it to the original latent dimensionality, providing a structured space in which collapse is explicitly penalized without imposing a bottleneck compression.
Operating in this projected latent space, a lightweight linear predictor $W_p$ learns to map the current state to the subsequent window: $\hat{\mathbf{S}}_{t+1} = W_p\,\pi_\phi(\mathbf{S}_t)$.
The predictor is intended as a deliberately restrictive linear transition head that pressures the encoder to place short-horizon dynamics into a state that remains easily predictable under a simple map.
Both the projector and predictor are strictly discarded following pretraining. In the \textsc{MAE} baseline, these latent-space heads are replaced by the standard forecasting \texttt{DecoderHead}; the encoder and downstream frozen-probe protocol remain unchanged.
Concretely, the \textsc{MAE} arm observes a 464-sample prefix, zero-pads it to length 512 for the backbone, and uses 48 learned query positions in a cross-attention decoder to predict the withheld 48-sample future under an L1 loss. This keeps the baseline architecturally close to the production encoder while making the supervisory signal fundamentally different from \textsc{LIN}: direct future-signal prediction rather than latent-state prediction.

\paragraph{Downstream probe.}
For nominal downstream evaluation, the pretrained encoder is frozen and only the mean-pooled final-MLP embedding $\mathbf{e}$ is exposed to a lightweight logistic-regression probe.
For a task with $K$ output classes, the probe is
\begin{equation}\label{eq:linear_probe}
  \hat{\mathbf{y}} = \mathrm{softmax}(W_{\mathrm{cls}} \mathbf{e} + b), \quad
  W_{\mathrm{cls}} \in \mathbb{R}^{K \times d}.
\end{equation}
This setup keeps the evaluation protocol faithful to the representation-learning objective: all task-specific adaptation is restricted to the linear head, while the encoder parameters remain fixed.
In the nominal TTL-TSFM lin variant, this probe is the sole downstream interface used for pathology discrimination, fall detection, and label-efficient HAR.

% sections/05_results.tex
% Experiments and results

\FloatBarrier
\section{Experiments and Results}\label{sec:results}
\subsection{Objective Comparison: Latent-State Prediction vs.\ Raw-Signal Forecasting}\label{subsec:objective_comparison}

\begin{table*}[!t]
\centering
\caption{Controlled comparison between the production latent world-model objective (\textsc{LIN}) and a causal raw-signal forecasting baseline (\textsc{MAE}, named for its mean-absolute-error loss) on the same hybrid encoder backbone. The comparison keeps the encoder, latent read-out, and downstream probe fixed, changing only the pretraining head and supervisory target: \textsc{LIN} predicts the latent state of the next window with projector + linear predictor + SIGReg, whereas \textsc{MAE} predicts the final 48 raw IMU samples from a 464-sample context via a cross-attention decoder. Bold indicates the better value in each row; OOD marks PAMAP2-based probes withheld from pretraining. Arrows indicate the preferred direction of each metric.}
\label{tab:main_results}
\scriptsize
\setlength{\tabcolsep}{4pt}
\resizebox{\textwidth}{!}{%
\begin{tabular}{llllll}
\toprule
\textbf{Axis} & \textbf{Eval.} & \textbf{Dataset} & \textbf{Primary metric} & \textbf{LIN} & \textbf{MAE} \\
\midrule
\multirow{6}{*}{Clinical discriminability}
  & E1 & WearGait-PD (PD vs HC) & AUC-ROC ($\uparrow$) & \textbf{0.756} & 0.631 \\
  & E1 & WearGait-PD (PD vs HC) & Macro-F1 ($\uparrow$) & \textbf{0.717} & 0.542 \\
  & E2 & WearGait-PD (5-cohort) & Macro-F1 ($\uparrow$) & \textbf{0.428} & 0.251 \\
  & E3 & Mobilise-D TVS (PD vs HA) & AUC-ROC ($\uparrow$) & \textbf{0.613} & 0.461 \\
  & E3 & Mobilise-D TVS (PD vs HA) & Binary F1 ($\uparrow$) & \textbf{0.634} & 0.526 \\
  & E5 & Voisard (7-condition) & Macro-F1 ($\uparrow$) & \textbf{0.280} & 0.162 \\
\midrule
\multirow{3}{*}{Data-efficiency and robustness}
  & E4 & LTMM (fall risk) & AUC-ROC ($\uparrow$) & \textbf{0.670} & 0.621 \\
  & E7 & Mobilise-D TVS + AWGN & Macro-F1 clean ($\uparrow$) & \textbf{0.319} & 0.204 \\
  & E7 & Mobilise-D TVS + AWGN & Macro-F1 @ 10 dB ($\uparrow$) & \textbf{0.284} & 0.213 \\
\midrule
\multirow{4}{*}{Generalisation and transfer}
  & E9 & SisFall SA$\rightarrow$SE & AUC-ROC ($\uparrow$) & \textbf{0.963} & 0.874 \\
  & E9 & SisFall SA$\rightarrow$SE & Binary F1 ($\uparrow$) & \textbf{0.846} & 0.682 \\
  & E10 & SisFall direction labels & Macro-F1 ($\uparrow$) & 0.536 & \textbf{0.565} \\
  & E11 & UCI HAR$\rightarrow$PAMAP2 (target OOD) & Macro-F1 ($\uparrow$) & 0.513 & \textbf{0.568} \\
\midrule
\multirow{5}{*}{Representation quality and structure}
  & E12 & WearGait-PD decoder probe & Val MAE ($\downarrow$) & 0.00260 & \textbf{0.00224} \\
  & E13 & WearGait-PD activations & Effective rank ($\uparrow$) & \textbf{59.1} & 33.0 \\
  & E13 & Mobilise-D TVS dynamical latents & Dynamics H-stat ($\uparrow$) & \textbf{15.26} & 12.08 \\
  & E13 & WearGait-PD trajectories & Latent straightness ($\uparrow$) & \textbf{-0.2973} & -0.3798 \\
  & E14 & UCI HAR forecast & Forecast MAE @ $h{=}48$ ($\downarrow$) & 0.02131 & \textbf{0.01869} \\
\bottomrule
\end{tabular}
}
\end{table*}

We evaluate \sonata through a deliberately broad but clinically grounded suite designed to test not only pathology discrimination, but also data efficiency, robustness, transfer, and latent structure. The main text keeps this coverage compact: it first states the evaluation logic, then reports two controlled comparisons---one isolating the pretraining objective and one isolating the predictor head. Full task definitions, datasets, protocols, and metric choices are deferred to Appendix~\ref{app:evaluation_protocol}.

\subsection{Evaluation Suite}
The suite is organized into four axes. Clinical discriminability probes whether the representation separates disease state, disease subtype, and prospective fall risk under frozen-encoder evaluation (E1--E5). Data-efficiency and robustness examine adaptation under sparse labels, injected sensor noise, and a partially supervised upper bound (E6--E8). Generalisation and transfer test cross-age fall detection, fall-direction sensitivity, and cross-dataset activity-transfer behaviour (E9--E11). Representation quality and structure then ask whether the learned state retains physically decodable information, avoids collapse, and supports short-horizon forecasting behaviour (E12--E14).

\begin{figure*}[!t]
\centering
\includegraphics[width=\textwidth]{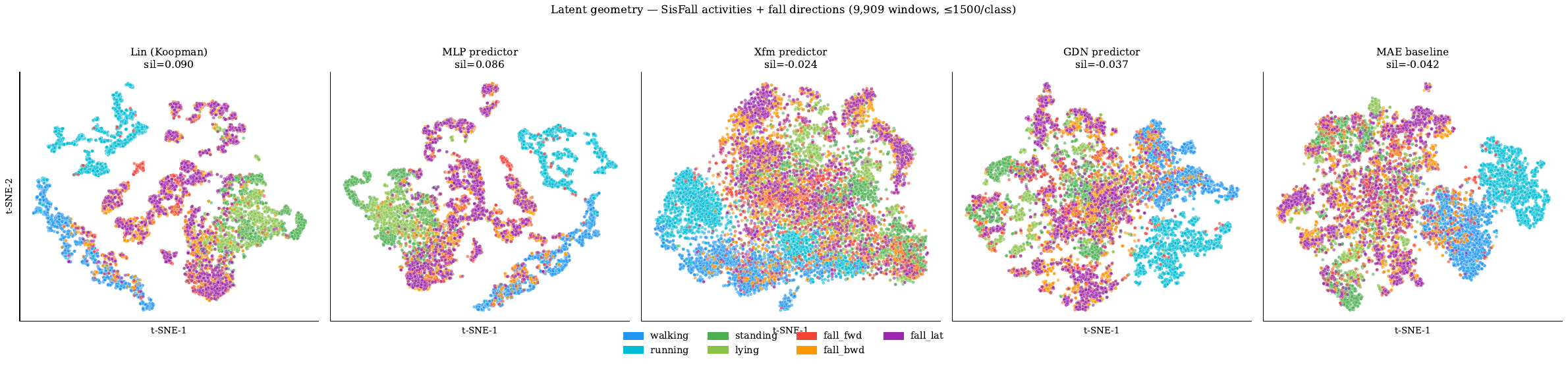}
\caption{Latent geometry on SisFall activities and fall directions, shown as two-dimensional t-SNE projections of the terminal latent state for the production \textsc{LIN} model, three alternative predictor heads, and the raw-signal forecasting \textsc{MAE} baseline. The visualization is designed to be read comparatively: while no model achieves clean class separation, the token-level world-model variants (\textsc{LIN} and MLP) preserve more coherent local clustering than the sequence-level predictors or the \textsc{MAE} baseline, mirroring the quantitative performance gaps in Table~\ref{tab:main_results} and Table~\ref{tab:predictor_ablation}. The residual overlap across all panels is itself informative, suggesting that SisFall imposes a meaningful separability ceiling under this windowing and labeling setup.}
\label{fig:sisfall_latent_geometry}
\end{figure*}

Table~\ref{tab:main_results} isolates the effect of the pretraining objective by holding the hybrid encoder and frozen-probe protocol fixed, while changing only the pretraining head and loss. \textsc{LIN} predicts the latent state of the next 512-step window via the projector--predictor stack and SIGReg; \textsc{MAE} predicts the final 48 raw IMU samples from a 464-sample observed prefix via the standard cross-attention forecasting decoder and an L1 loss. The resulting pattern is mixed rather than one-sided. \textsc{LIN} leads on the clinically central and robustness-oriented family of probes: both WearGait PD-versus-control endpoints, the harder five-cohort WearGait task, the Mobilise-D PD-versus-healthy comparison, LTMM fall-risk discrimination, cross-age fall detection, both noise rows, and the low-label PAMAP2 adaptation curve. \textsc{MAE}, by contrast, is strongest on the rows most closely coupled to waveform fidelity or local motion templates: the post-hoc decoder probe, the single-horizon forecast MAE, SisFall fall-subtype recognition (0.565 vs.\ 0.536), and UCI-to-PAMAP2 transfer (0.568 vs.\ 0.513). The most plausible interpretation is therefore not that one objective is universally superior, but that they preserve different aspects of the signal: \textsc{MAE} stays closer to the sensor waveform, whereas \textsc{LIN} more aggressively organizes latent factors that survive frozen probing under cohort heterogeneity and sensor corruption.

This distinction matters because several of the hardest rows in the table have intentionally modest absolute scores. We do not read those values as evidence of clinical readiness, nor as grounds to dismiss the comparison; instead, they should be interpreted as comparative evidence under a difficult protocol. On heterogeneous subject-level problems such as PD-vs-HA, multicohort gait discrimination, or cross-dataset transfer, absolute performance is compressed by label noise, cohort overlap, and domain shift. In that setting, the relevant signal is the recurrence of the same ordering across related stress tests. \textsc{LIN}'s gains are therefore meaningful less because any single number is large in isolation, and more because the same objective advantage reappears across pathology discrimination, fall-risk ranking, robustness to injected noise, and low-label adaptation. By the same logic, the MAE-favorable rows are informative rather than inconvenient: they identify the settings in which literal signal reconstruction remains a strong proxy for downstream usefulness.

The structural diagnostics reinforce this split. \textsc{LIN} produces nearly twice the effective rank of \textsc{MAE} (59.1 vs.\ 33.0)\footnote{Effective rank is the exponential of the Shannon entropy of the normalised singular value distribution of the activation matrix; it measures how many representation dimensions are meaningfully used~\citep{Roy2007EffectiveRank}.}, a stronger latent-dynamics separation signal (H = 15.26 vs.\ 12.08), and straighter latent trajectories ($-0.297$ vs.\ $-0.380$), all consistent with the world-model hypothesis developed in Section~\ref{sec:architecture}. \Cref{fig:sisfall_latent_geometry} renders the same ordering geometrically. The t-SNE projections show that token-level latent prediction (\textsc{LIN} and MLP) yields more coherent local islands than either sequence-level prediction (Transformer and GDN) or raw-signal forecasting (\textsc{MAE}), which appear progressively more entangled. Crucially, we do not interpret this figure as evidence of clean, absolute class clustering---all five panels retain substantial overlap---but that is precisely the point. For a heterogeneous, noisy fall dataset such as SisFall, the residual mixing points to a nontrivial data ceiling. This split follows directly from the objective-level distinction developed in \cref{sec:architecture}: \textsc{MAE} rewards intra-window waveform fidelity, allocating encoder capacity to high-frequency temporal structure that need not persist across windows; \textsc{LIN} rewards cross-window latent predictability, allocating that capacity instead to structure that does. The empirical pattern is exactly what this distinction predicts. The practical bottom line is therefore task-aligned rather than universal: if the goal is short-horizon waveform fidelity or signal-faithful decoding, \textsc{MAE} remains a sensible objective; if the goal is stable clinical state separation under heterogeneity, limited labels, and sensor corruption, \textsc{LIN} is the better-matched choice.

\subsection{Predictor Ablation: Why a Linear Head Suffices}\label{subsec:predictor_ablation}

\begin{table*}[!t]
\centering
\caption{Predictor ablation under a fixed JEPA encoder. Only the predictor head varies; the encoder, projector, data pipeline, and training schedule are shared. Bold marks the best value per row. PAMAP2 is fully withheld from pretraining (OOD). GDN training diverged at epoch~4; results reflect this early checkpoint. Arrows indicate the preferred direction of each metric.}
\label{tab:predictor_ablation}
\scriptsize
\setlength{\tabcolsep}{4pt}
\begin{tabular}{lllllll}
\toprule
\textbf{Axis} & \textbf{Eval.} & \textbf{Primary metric} & \textbf{Linear} & \textbf{MLP} & \textbf{Transformer} & \textbf{GDN\textsuperscript{$\dagger$}} \\
\midrule
\multirow{2}{*}{Clinical}
  & WearGait 5-cohort & Macro-F1 ($\uparrow$) & \textbf{0.428} & 0.387 & 0.158 & 0.209 \\
  & Voisard 7-cohort & Macro-F1 ($\uparrow$) & \textbf{0.280} & 0.247 & 0.173 & 0.175 \\
\midrule
\multirow{7}{*}{OOD / robustness}
  & PAMAP2 Acc. @ 1\% & Acc. ($\uparrow$) & \textbf{67.1 $\pm$ 5.3} & 63.1 $\pm$ 6.1 & 41.3 $\pm$ 4.0 & 47.4 $\pm$ 4.9 \\
  & PAMAP2 Acc. @ 5\% & Acc. ($\uparrow$) & \textbf{78.7 $\pm$ 2.1} & 76.2 $\pm$ 2.7 & 54.3 $\pm$ 3.0 & 62.4 $\pm$ 4.3 \\
  & PAMAP2 Acc. @ 10\% & Acc. ($\uparrow$) & \textbf{80.1 $\pm$ 2.2} & 78.3 $\pm$ 1.8 & 59.5 $\pm$ 2.2 & 66.8 $\pm$ 2.6 \\
  & PAMAP2 Acc. @ 20\% & Acc. ($\uparrow$) & \textbf{81.9 $\pm$ 1.7} & 80.4 $\pm$ 1.5 & 62.8 $\pm$ 1.8 & 69.6 $\pm$ 2.0 \\
  & PAMAP2 Acc. @ 50\% & Acc. ($\uparrow$) & \textbf{83.0 $\pm$ 1.0} & 82.4 $\pm$ 1.2 & 66.3 $\pm$ 1.0 & 72.0 $\pm$ 0.8 \\
  & TVS noise clean & Macro-F1 ($\uparrow$) & \textbf{0.319} & 0.311 & 0.096 & 0.181 \\
  & TVS noise @ 10 dB & Macro-F1 ($\uparrow$) & 0.284 & \textbf{0.309} & 0.061 & 0.041 \\
\midrule
\multirow{1}{*}{Transfer}
  & UCI $\rightarrow$ PAMAP2 & Macro-F1 ($\uparrow$) & 0.513 & 0.423 & 0.420 & \textbf{0.562} \\
\midrule
\multirow{2}{*}{Structure}
  & Effective rank & Rank ($\uparrow$) & 59.1 & 57.8 & 65.0 & \textbf{72.5} \\
  & Latent straightness & Mean cosine ($\uparrow$) & $-$0.297 & \textbf{$-$0.267} & $-$0.444 & $-$0.400 \\
\bottomrule
\end{tabular}

\vspace{2pt}
{\raggedright\footnotesize\textsuperscript{$\dagger$}GDN training diverged at epoch~4/30; all GDN entries reflect this early checkpoint.\par}
\end{table*}

Having established the value of the latent world-model objective in Section~\ref{subsec:objective_comparison}, we now ask a complementary question: given that objective, does the \emph{capacity} of the predictor head matter? Table~\ref{tab:predictor_ablation} holds the encoder, projector, data pipeline, and training schedule fixed, varying only the predictor among four architectures that span two orthogonal axes of expressivity: nonlinearity and temporal scope.

\textbf{Linear} ($W_p \in \mathbb{R}^{128 \times 128}$, no bias, no nonlinearity) operates on the projected terminal token $\pi(\mathbf{S}_T)$ alone and is the production predictor described in Section~\ref{sec:architecture}. \textbf{MLP} ($128 \!\to\! 512 \!\to\! 128$ with LayerNorm and GELU) also operates on $\pi(\mathbf{S}_T)$, adding a single hidden nonlinearity while remaining token-level. \textbf{Transformer} (two-layer bidirectional \texttt{TransformerEncoder}, $d{=}128$, $h{=}4$, $d_\text{ff}{=}512$, with learned positional embedding) and \textbf{GDN} (single-block Gated DeltaNet, $d{=}128$, $h{=}4$, $\mathrm{expand\_v}{=}2$) both promote the predictor to a sequence-level module: the projector is applied position-wise to the full encoder output $(B, 512, 128)$, and the MSE loss is computed over all positions rather than only the terminal token. All four arms share the same projector, training loss, and downstream frozen-probe protocol; the only degree of freedom is the predictor itself.

The clearest signal comes from the withheld PAMAP2 label-efficiency rows, where each entry reports mean accuracy $\pm$ standard deviation over five random subsampling seeds on a fixed held-out test set of two subjects (956 windows). The linear predictor is best at every label fraction from 1\% to 50\%, and the advantage is largest in the most data-scarce regime (67.1\% vs.\ 63.1\% at 1\%; +25.8 pp over the Transformer). The same ordering holds on the harder multicohort clinical probes: the linear head leads on both WearGait 5-cohort (+0.041 over MLP, +0.270 over Transformer) and Voisard 7-cohort classification. Under clean noise conditions the linear head is again strongest; the MLP is marginally better at 10\,dB, suggesting that mild predictor flexibility is not inherently harmful but does not translate into a systematic advantage.

The result is not uniformly one-sided. GDN achieves the highest UCI$\rightarrow$PAMAP2 transfer score (0.562 vs.\ 0.513 for linear), but training diverged at epoch~4 and this entry reflects an early checkpoint rather than a converged model; the high effective rank (72.5) paired with poor latent straightness ($-0.400$, second only to the Transformer's $-0.444$) suggests an unstable, high-entropy representation rather than a structured one. The Transformer head is the most informative negative result: despite substantially greater capacity, it produces the worst multicohort clinical scores (0.158 WearGait, 0.173 Voisard) and roughly halves PAMAP2 accuracy relative to the linear head at every label fraction (41.3\% vs.\ 67.1\% at 1\%), consistent with the predictor absorbing temporal dynamics that the encoder should internalize for frozen-probe evaluation. The more precise discriminant is temporal scope rather than capacity alone: both token-level predictors (Linear and MLP, operating on the terminal token $\pi(\mathbf{S}_T)$) lead clinical probes, while the Transformer promotes the predictor to a sequence-level module over the full 512-step encoder output and produces the worst clinical scores. The MLP is the critical partial control: it adds nonlinearity without temporal scope and incurs only a modest clinical penalty (WearGait 5-cohort 0.387 vs.\ 0.428 for linear)---a result inconsistent with a pure optimisation-difficulty account, which predicts smooth degradation with capacity rather than a categorical drop at the token-to-sequence boundary. The practical conclusion is that a deliberately simple linear transition head is the most reliable choice across sparse-label transfer, heterogeneous clinical separation, and training stability---not because expressive heads cannot learn, but because they remove the inductive pressure that makes the encoder's frozen representation clinically useful.

% sections/06_ablation_studies.tex
% Ablation Studies

\section{Ablation Studies}\label{sec:ablations}

\begin{table*}[!t]
  \centering
  \caption{%
    Consolidated ablation summary. Each row changes one design decision
    relative to production; all other hyperparameters are held fixed.
    Predictor-head variants are omitted (see \cref{tab:predictor_ablation}).
    \textbf{Bold}: production value, or a strict improvement over it on
    that column.
  }
  \label{tab:main_ablation_summary}
  \small
  \setlength{\tabcolsep}{5pt}
  \begin{tabular}{llcccc}
    \toprule
    \multirow{2}{*}{\textbf{Variant}} &
    \multirow{2}{*}{\textbf{Change}} &
    \textbf{PD/HC} &
    \textbf{WG\,5-coh.} &
    \textbf{Faller} &
    \textbf{Fall det.} \\
    & & AUC $\uparrow$ & F1 $\uparrow$ & AUC $\uparrow$ & AUC $\uparrow$ \\
    \midrule
    \textbf{LIN-0.2 (prod)} & --- & \textbf{0.757} & \textbf{0.428} & \textbf{0.670} & \textbf{0.963} \\
    \midrule
    \multicolumn{6}{l}{\textit{SIGReg weight}} \\
    \quad $\lambda{=}0$    & SIGReg off   & 0.581          & 0.114          & 0.568          & 0.750 \\
    \quad $\lambda{=}0.05$ & Lighter reg. & 0.746          & 0.388          & \textbf{0.732} & \textbf{0.983} \\
    \quad $\lambda{=}0.1$  & Intermediate & \textbf{0.785} & 0.423          & 0.611          & \textbf{0.969} \\
    \quad $\lambda{=}0.5$  & Over-reg.    & 0.686          & 0.306          & 0.659          & 0.854 \\
    \midrule
    \multicolumn{6}{l}{\textit{Eigenvalue range}} \\
    \quad Restricted eigsp. & $\beta\in(0,1)$ & 0.743 & 0.396 & 0.665 & \textbf{0.966} \\
    \midrule
    MAE (baseline) & Signal forecast & 0.631 & 0.251 & 0.621 & 0.874 \\
    \bottomrule
  \end{tabular}
\end{table*}

% ─────────────────────────────────────────────────────────────────────────────
\subsection{Setup and Common Protocol}\label{subsec:ablations_setup}
% ─────────────────────────────────────────────────────────────────────────────

All ablations share a fixed encoder backbone ($\approx$3.60\,M encoder
parameters, 12 blocks, $d_{\mathrm{model}}{=}128$), the same nine-dataset
lower-back IMU corpus, and the same 30-epoch warmup--steady--decay schedule
with $\mathrm{lr}{=}10^{-3}$. Only the named hyperparameter varies; all
others are held at their production values. Predictor-head results are
reported separately in \cref{subsec:predictor_ablation} and are not
re-tabulated here. All rows are sourced from the same
\texttt{tier1\_extended\_v3} evaluation suite as \cref{tab:main_results},
using the four headline probes: WearGait PD/HC AUC, WearGait five-cohort
macro-F1, LTMM faller AUC, and SisFall SA$\to$SE fall-detection AUC.

% ─────────────────────────────────────────────────────────────────────────────
\subsection{SIGReg Anti-Collapse Regularisation}\label{subsec:sigreg_ablation}
% ─────────────────────────────────────────────────────────────────────────────

Without additional regularisation, JEPA objectives with symmetric gradient
flow are susceptible to representational collapse: the encoder maps all
inputs to a constant vector that trivially minimises the prediction term in
\cref{eq:lwm_loss}. \sonata uses the Sketched Isotropic Gaussian
Regulariser (SIGReg) of \citet{Balestriero2025LeJEPA}, which approximates a
maximum-entropy penalty via $M{=}1024$ random Gaussian projections and
$K{=}17$ quadrature knots, enforcing embedding isotropy without
stop-gradients or asymmetric target encoders. We sweep $\lambda \in
\{0,\,0.05,\,0.1,\,0.2,\,0.5\}$; full results are in
\cref{tab:main_ablation_summary}.

\paragraph{SIGReg is load-bearing.}
Disabling SIGReg ($\lambda{=}0$) causes catastrophic collapse: WG\,5-cohort
F1 falls to 0.114, below the MAE baseline (0.251), and PD/HC AUC drops to
0.58---near-random for a two-class 2:1 split. The validation loss of 0.073
confirms trivial minimisation via constant representations, yet this failure
is not detectable from training loss alone; it surfaces only in downstream
probe evaluation.

\paragraph{$\lambda$ admits a flat optimum across $[0.1,\,0.2]$.}
$\lambda{=}0.2$ is the nominal WG\,5-cohort optimum, with $\lambda{=}0.1$
statistically tied ($\Delta{=}0.005$) and leading on PD/HC AUC
($+$2.9~pts). $\lambda{=}0.05$ is Pareto-optimal on fall tasks---LTMM
$+$6.2~pts, SisFall $+$2.0~pts---at the cost of $-$4.0~pts on
WG\,5-cohort, offering a principled alternative for deployments focused
exclusively on fall risk. Beyond $\lambda{\approx}0.2$, the SIGReg penalty
swamps the prediction objective: $\lambda{=}0.5$ loses $-$12.1~pts on
WG\,5-coh and $-$7.1~pts on PD/HC.

% ─────────────────────────────────────────────────────────────────────────────
\subsection{GatedDeltaNet Eigenvalue Range}\label{subsec:eigval_ablation}
% ─────────────────────────────────────────────────────────────────────────────

The GatedDeltaNet recurrence (\cref{eq:gdn_recurrence}) includes a gating
scalar $\beta_t = \sigma(W_\beta \mathbf{x}_t)$. The production
configuration permits $\beta_t \in (0, 2)$, so the effective eigenvalue
$1 - \beta_t \in (-1, 1)$, admitting oscillatory modes. Constraining
$\beta_t \in (0, 1)$ restricts eigenvalues to $(0, 1)$, limiting dynamics
to pure exponential decay. \citet{Grazzi2025StateTracking} prove that this
restriction prevents linear recurrences from solving state-tracking tasks
that require periodic or sign-alternating dynamics.

\paragraph{Train--probe divergence.}
Restricting the eigenspectrum \emph{improves} pretraining validation loss
(0.030 vs.\ 0.034) while \emph{degrading} clinical probes: WG\,5-cohort
$-$3.2~pts, PD/HC $-$1.4~pts. Monotone-decay dynamics are more predictable
under the JEPA prediction loss but produce a shallower encoder.

\paragraph{Gait requires oscillatory modes.}
Gait is a multi-scale periodic process---stride cycles at $\approx$1\,Hz,
rotational dynamics at 2--4\,Hz, and stride-to-stride variability at
0.1--0.25\,Hz---whose latent representation requires sign-alternating
recurrent components that exponential decay cannot express. The production
model actively exploits the extended eigenspectrum: gate statistics at the
final recurrent layer place the mean eigenvalue at $\approx 0$ with
substantial probability mass below zero, yielding an effective memory
horizon of $\approx$4.1\,s---spanning 3--4 complete stride cycles---an
emergent timescale shaped by training on gait data rather than imposed as
a prior. Restricting to positive eigenvalues removes this capacity and,
as a further geometric consequence, produces more curved latent trajectories
($-$0.327 vs.\ $-$0.297 for production under a mean-pool temporal
straightening probe), consistent with \citet{Maes2026LeWM}'s finding that
JEPA with SIGReg induces latent straightening as a world-model property.
Fall detection is insensitive to the constraint (0.966 vs.\ 0.963):
acute fall events are dominated by impulsive transients that
exponential-decay modes capture adequately.

% ─────────────────────────────────────────────────────────────────────────────
\subsection{Production Configuration Rationale}\label{subsec:prod_rationale}
% ─────────────────────────────────────────────────────────────────────────────

The production model is selected on WG\,5-cohort macro-F1 while remaining
Pareto-undominated on the other three probes. The linear predictor
(\cref{tab:predictor_ablation}) is the primary structural driver; $\lambda
{=}0.2$ is the nominal WG\,5-cohort optimum, with $\lambda{=}0.1$
statistically tied and marginally stronger on PD/HC, giving a flat
operating region across $[0.1,\,0.2]$; and the full eigenspectrum
contributes $+$3.2~pts on WG\,5-cohort at the cost of marginally higher
pretraining loss, consistent with the theoretical necessity of oscillatory
modes for periodic gait dynamics. Predictor-head variants are reported in
\cref{tab:predictor_ablation}.

% sections/07_discussion.tex
% Discussion

\section{Discussion}\label{sec:discussion}

The results bear out the consequences of this design discipline.
A representation learned under a latent predictive objective produces, in this evaluation, richer latent structure and stronger clinical discriminability than the same architecture trained under a more permissive signal-reconstruction objective on the same data.
The implications extend from immediate deployment properties to a longer-horizon clinical monitoring use case that the architecture is particularly well placed to support.

Matching capacity to the kinematic degrees of freedom of pathological gait rather than to dataset volume produces a model that is both fitted to the signal and feasible to deploy---not because these goals are in tension and a compromise was struck, but because in the clinical data regime they point to the same scale.
At 3.77 million parameters, \sonata is compatible with inference on embedded processors within wearable IMU systems, removing the need to stream raw kinematics to external endpoints.
This matters beyond engineering convenience.
Gait signatures recorded at clinical-grade sampling rates are sensitive data: they carry identifying biometric information and, in continuous form, can expose behavioural patterns well beyond locomotion.
On-device inference removes the transmission pathway; only compressed, task-specific semantic outputs---a fall probability estimate, a daily gait health aggregate---leave the device, reducing the data-governance overhead associated with continuous off-device transmission and simplifying the data-handling requirements of multi-site studies.
The compressed latent state can further be computed at stride-scale latency, enabling closed-loop applications---freezing-of-gait cueing, prospective fall alerting---that are sensitive to network delay in ambulatory settings.

Capacity discipline also has consequences for model auditability.
A short computational graph with a structured, physics-motivated latent space is more amenable to the systematic sensitivity analysis that regulated deployment pathways demand.
The failure mode space is more tractable: when the model's latent representation is organised around biomechanical dynamics rather than sensor statistics, unexpected predictions are more readily traced to clinically interpretable causes.
For medical device qualification, this property is not peripheral: it bounds the space of expected behaviour and simplifies the validation evidence required to establish trust.
The structural diagnostics---higher effective rank, straighter latent trajectories, and stronger dynamical separation---are consistent with the hypothesis that the encoder has internalised a compact model of kinematic dynamics rather than memorising dataset-specific sensor statistics.
Taken together, the results suggest that pretraining objective choice is task-aligned rather than universal: latent prediction is better matched to clinical state separation under heterogeneity, whereas raw forecasting remains advantageous for waveform-faithful reconstruction tasks.

\subsection{Limitations}

The present results establish representation quality under controlled frozen-probe evaluation; they are not claims of clinical deployment readiness, and prospective validation on longitudinal endpoints remains the subject of ongoing work.

The principal limitation of the present study is the narrow admissible sensing regime. Pretraining is restricted to clinically meaningful six-axis inertial measurements with preserved accelerometer and gyroscope channels, from trunk-centred placements where translational and rotational kinematics remain directly interpretable. This imposes a real data ceiling and excludes many otherwise attractive wearable datasets, so the present results should be interpreted as evidence for representation quality in a high-fidelity clinical regime rather than as a demonstration of broad coverage across arbitrary wearable setups.

The most consequential implication, however, concerns how the representation can be used once deployed longitudinally.
The prevailing approach to tracking disease progression in neurology relies on episodic clinical rating scales that are infrequent, subject to rater variability, and insensitive to the day-to-day fluctuations that affect patient quality of life.
Wearable digital biomarkers address the frequency problem, but population-level thresholds remain a blunt instrument for individual monitoring: inter-subject variability in healthy gait is large enough that a patient who has deteriorated meaningfully from their own baseline may still fall within the normal range for their demographic group.
The world-model framing offers a principled route through this.
Because SIGReg pressure encourages the latent space to organise around slowly varying physiological dynamics rather than sensor-level idiosyncrasies, a patient's trajectory through that space over time encodes a personal kinematic fingerprint.
Deviation from that personal trajectory---rather than from a population norm---becomes the monitoring signal, factoring out inter-subject variability as baseline structure.
A patient whose latent trajectory consistently diverges from the model's learned expectations may be exhibiting kinematics the model has not encountered before, making such divergence a potentially sensitive early indicator of disease progression without requiring explicit re-training or fine-tuning.
If confirmed prospectively, this shifts the basis of longitudinal assessment from population rank to individual dynamical change, a property that is increasingly necessary for detecting subtle treatment effects in early-stage disease cohorts.
The present paper demonstrates the representational properties---structured latent geometry, clinical discriminability across cohorts, robustness to noise---that would support this use case; its prospective validation remains an explicit target of future work.

% sections/06_conclusion.tex
% Conclusion (merged Discussion + Conclusion)

\section{Conclusion}\label{sec:conclusion}

Clinical-grade IMU data are small in scale and hard to acquire, and \sonata is designed from that reality upward.
Its sensor constraints, lightweight backbone, latent world-model objective, and physics-motivated regularisation are all answers to the same question: how do you learn representations that remain clinically reliable when data are limited but measurement quality is high?
To the best of our knowledge, this is also the first application of world-model pretraining to IMU kinematics---a contribution that is not only architectural but conceptual, reframing the learning problem from reconstructing raw sensor traces to predicting biomechanical state.
As the results suggest, that distinction carries through to representation quality: the latent space is better organised, clinical discriminability is stronger, and robustness to noise degradation is improved relative to a forecasting objective trained on the same corpus.

\paragraph{Outlook.}
The immediate next step is to relax the current single-lumbar assumption and support variable sensor configurations, including multi-IMU setups when clinically available, while retaining both accelerometer and gyroscope channels as the minimal clinically informative sensing pair. Doing so is not simply a matter of adding more files: the model must learn to handle differences in sensor geometry, body-site semantics, and inter-sensor coordination without discarding the coupled translational--rotational structure that carries much of the clinically relevant variability. If that extension is successful, it should enlarge the admissible pretraining corpus and allow the model to learn coordinated whole-body dynamics across placements---an important step toward richer kinematic representations for more complex movement disorders.
At sufficient data scale, we expect the same world-model objective to support more foundation-model-like behaviour: broader cross-cohort transfer, stronger zero-shot task generalisation, and greater sensitivity to personalised longitudinal endpoints.
A general-purpose kinematic world model that transfers robustly across neurological conditions, sensor configurations, and acquisition protocols would represent a meaningful advance for clinical trial design in neurology, where the bottleneck is increasingly statistical power rather than technology.
Reaching that regime is a central aim of the programme.

% -----------------------------------------------------------------------
\bibliographystyle{unsrtnat}
\bibliography{bib/references}

% -----------------------------------------------------------------------
% Draft Outlines
% -----------------------------------------------------------------------
\inmode{Draft}{
  \clearpage
  \begingroup
  \color{blue}
  \section*{Paper Draft Checklist}
  
  \subsection*{\texttt{v1.0}}
  \subsubsection*{Ablations}
  \begin{itemize}[label=\textcolor{blue}{$\square$}]
    \item Projector: 128 vs 64 dim compression\vspace{0.5em}
    \item Justify adopting project MLP as in Laya\vspace{0.5em}
    \item Turn off SIGReg\vspace{0.5em}
  \end{itemize}
  \subsubsection*{Narrative}
  \begin{itemize}[label=\textcolor{blue}{$\square$}]
    \item Explain 3:1 block structure, varying from Reverso, and flipping order of ops\vspace{0.5em}
    \item Justify window size (Reverso) and 100 Hz Sampling Rate\vspace{0.5em}
    \item Result: physics understanding; outlook: first step towards FM (multi-IMU expansion)\vspace{0.5em}
  \end{itemize}
  \subsubsection*{Experiments \& Results}
  \begin{itemize}[label=\textcolor{blue}{$\square$}]
    \item Probe eigen-expansion to [-1,1] in beta for GDN.\vspace{0.5em}
    \item $\lambda$ hyperparam tuning\vspace{0.5em}
    \item Comparison with big-tech TSFM\vspace{0.5em}
    \item Swap out bidirectional for causal attention in predictor layer\vspace{0.5em}
    \item Prediction horizon\vspace{0.5em}
  \end{itemize}
  \subsubsection*{Figs}
  \begin{itemize}[label=\textcolor{blue}{$\square$}]
    \item \vspace{0.5em}
    \item \vspace{0.5em}
  \end{itemize}

  \subsection*{\texttt{v1.1}}
  \subsubsection*{Experiments \& Results}
  \begin{itemize}[label=\textcolor{blue}{$\square$}]
    \item Dissect linear prediction to position Koopman as legitimate claim $\rightarrow$ \emph{canonical transform of single-sensor human biomechanics}\vspace{0.5em}
    \item Revisit window size and sampling rate\vspace{0.5em}
    \item Appraise multi-scale loss LWM loss\vspace{0.5em}
    \item Understand what in the reps drives better clinical F1\vspace{0.5em}
  \end{itemize}

  \subsection*{\texttt{v2.0}}
  \subsubsection*{Expansion Stack}
  \begin{itemize}[label=\textcolor{blue}{$\square$}]
    \item Positional encoding of IMU sensors (admit arbitrary datasets)\vspace{0.5em}
    \item Training and evals corpus expansion, targeting FM capabilities\vspace{0.5em}
  \end{itemize}
  \endgroup
  \clearpage
}

% -----------------------------------------------------------------------
% Appendix
% -----------------------------------------------------------------------
\appendix

\inmode{Draft}{
% sections/A2_augmentation_ablations.tex
% Appendix B: Data Processing and Augmentations

\section{Data Processing and Augmentations}\label{app:augmentation}

\todo{Move augmentation ablation tables, per-augmentation math, and synthetic data generator specification here from 03\_data.tex.}

% sections/A1_data_details.tex
% Appendix A: Pretraining Corpus Details

\section{Pretraining Corpus Details}\label{app:data_details}

\todo{Preprocessing details to be ported from \texttt{sections/03\_data.tex}
once main-text experiments settle: per-dataset provenance and subject
demographics, ISB axis rotation matrices, resampling parameters, dual-stride
windowing equations, fixed-range scaling rationale, HDF5 schema and chunk
layout, and post-write verification checks.}

}
% sections/A4_evaluation_protocol.tex
% Evaluation protocol appendix

\section{Evaluation Protocol}\label{app:evaluation_protocol}

\begin{table*}[!t]
\centering
\caption{Evaluation battery used to assess frozen encoder representations. LOOCV denotes leave-one-subject-out cross-validation; AWGN denotes additive white Gaussian noise. Scope indicates whether a probe is planned for all predictor arms or only for the primary shared-backbone \textsc{LIN}/\textsc{MAE} comparison, where \textsc{MAE} denotes the raw-signal forecasting baseline trained with mean absolute error.}
\label{tab:appendix_eval_battery}
\scriptsize
\setlength{\tabcolsep}{4pt}
\resizebox{\textwidth}{!}{%
\begin{tabular}{cllllp{1.8cm}}
\toprule
\textbf{ID} & \textbf{Probe} & \textbf{Dataset} & \textbf{Protocol} & \textbf{Primary metric} & \textbf{Scope} \\
\midrule
E1 & PD vs.\ HC discrimination & WearGait-PD & 5-fold CV & AUC-ROC / Binary F1 & all \\
E2 & Five-cohort gait subtype discrimination & WearGait-PD & 5-fold CV & Macro-F1 & all \\
E3 & PD vs.\ HA discrimination & Mobilise-D TVS & LOOCV & AUC-ROC & all \\
E4 & Prospective fall-risk discrimination & LTMM & 5-fold CV & AUC-ROC & all \\
E5 & Seven-condition gait discrimination & Voisard & LOOCV & Macro-F1 & all \\
E6 & Label efficiency & PAMAP2 (OOD) & fixed split + 5 seeds & Accuracy @ \{1, 5, 10, 20, 50\}\% & predictor ablation \\
E7 & Noise robustness & Mobilise-D TVS & LOOCV + AWGN & Macro-F1 (clean, 10 dB) & all \\
E8 & Supervised upper bound & Mobilise-D TVS & LOOCV finetune & Macro-F1 & planned \\
E9 & Cross-age fall detection & SisFall SA$\rightarrow$SE & fixed split & AUC-ROC / Binary F1 & all \\
E10 & Fall direction classification & SisFall & LOOCV & Macro-F1 & all \\
E11 & Cross-dataset HAR transfer & UCI HAR$\rightarrow$PAMAP2 (target OOD) & fixed zero-shot & Macro-F1 & all \\
E12 & Post-hoc physical decoder & WearGait-PD & 80/20 split & Val MAE & all \\
E13 & Structural diagnostics & WearGait-PD / Mobilise-D TVS & activation sweep & Rank / dynamics H / straightness & all \\
E14 & Horizon forecast & UCI HAR & 80/20 split & Forecast MAE @ $h{=}48$ & all \\
\bottomrule
\end{tabular}
}
\end{table*}

All evaluations operate on a frozen encoder unless explicitly noted. During pretraining, the objective is defined on the terminal recurrent readout $\mathbf{S}_T$ from \cref{eq:latent_readout}. All downstream probes, however, use the mean-pooled final-MLP embedding $\mathbf{e}$ from \cref{eq:downstream_embedding}; the terminal recurrent state is not exposed directly at evaluation time. For the primary comparison, both \textsc{LIN} and \textsc{MAE} use the same encoder and the same downstream frozen-probe interface; they differ only in pretraining head and supervisory target. \textsc{LIN} predicts the latent state of the next window with a projector, linear predictor, and SIGReg regularizer, whereas \textsc{MAE} predicts the final 48 raw samples from a 464-sample observed prefix through the standard cross-attention forecasting decoder under an L1 loss. Unless a dataset-prescribed split is required, subject-level probes use leave-one-subject-out or subject-stratified cross-validation; window-level probes are reserved for tasks where temporal resolution is part of the endpoint itself, such as fall detection or forecasting. Only PAMAP2 is fully withheld from pretraining, so the rows touching PAMAP2 provide the clearest OOD evidence in the first-pass comparison.

\subsection{Datasets and Shared Evaluation Setup}
The battery spans six dataset families chosen to cover pathology discrimination, fall risk, activity transfer, and physical decoding. Mobilise-D TVS provides the primary multi-cohort clinical benchmark and the noise-robustness / supervised-ceiling evaluations. WearGait-PD supports both binary and fine-grained Parkinsonian discrimination, as well as post-hoc physical decoding. LTMM contributes prospective fall-risk prediction from laboratory walking trials. Voisard broadens the pathology landscape to seven clinical gait conditions. SisFall enables both cross-age fall detection and fall-direction classification. PAMAP2~\citep{Reiss2012PAMAP2} and UCI HAR support label-efficiency and cross-dataset transfer analyses.

All inputs are harmonized to a common representation: 6-axis trunk or waist IMU signals, resampled or downsampled to 100\,Hz, windowed to 512 samples (5.12\,s), and scaled using fixed physical sensor ranges rather than per-window normalization. This standardization keeps the evaluation focused on representational quality rather than dataset-specific preprocessing variance.

\subsection{Clinical Discriminability: E1--E5}
\paragraph{E1: WearGait-PD binary discrimination.}
This probe tests whether frozen embeddings separate Parkinson's disease from healthy controls under 5-fold subject-stratified cross-validation. The primary metric is AUC-ROC, with binary F1 and accuracy as supporting measures. A paired 6-axis versus 3-axis analysis isolates the contribution of gyroscope information.

\paragraph{E2: WearGait-PD five-cohort discrimination.}
This extends the previous task to finer-grained cohort separation, testing whether the representation captures intra-pathology structure rather than only broad disease-control differences. Macro-F1 is the primary metric because class balance is uneven and per-class performance matters.

\paragraph{E3: Mobilise-D PD versus healthy adults.}
This is a cross-dataset replication of Parkinsonian discrimination under a standardized laboratory protocol. LOOCV is used because of the modest cohort size, and AUC-ROC is again the headline metric.

\paragraph{E4: LTMM prospective fall risk.}
Unlike the pathology tasks, this probe asks whether a single laboratory gait assessment contains latent evidence of future fall risk. Binary F1 and balanced accuracy are emphasized because class balance and recall of future fallers are clinically meaningful.

\paragraph{E5: Voisard seven-condition discrimination.}
This is the broadest pathology benchmark in the battery, spanning neurological, orthopaedic, and healthy cohorts. Strong macro-F1 here would support the claim that the latent space captures clinically general gait structure rather than a single disease signature.

\subsection{Data-Efficiency and Robustness: E6--E8}
\paragraph{E6: Label efficiency on PAMAP2.}
The frozen encoder is paired with logistic-regression probes trained on progressively smaller fractions of labeled data. At each fraction, $x$\% of training windows per class are subsampled, the probe is fit, and accuracy is evaluated on a fixed held-out test set of two subjects (956 windows). This procedure is repeated five times with different random seeds; the reported $\pm$ values are standard deviations over those five runs. We report accuracy at 1\%, 5\%, 10\%, 20\%, and 50\% label fractions directly, rather than collapsing these points into a single AUC-LEC summary, because these are the exact completed outputs available for the first sweep.

\paragraph{E7: Noise robustness on Mobilise-D TVS.}
Additive white Gaussian noise is injected at test time while probes remain trained on clean data. The present comparison reports macro-F1 in the clean condition and at 10\,dB directly, rather than a derived degradation statistic, so that the appendix matches the completed table entries.

\paragraph{E8: Supervised upper bound.}
The final backbone block and a lightweight head are partially fine-tuned under LOOCV to estimate the remaining headroom above the frozen representation. The gap between this ceiling and the frozen-probe result serves as a compact measure of how much task-relevant structure pretraining has already captured.

\subsection{Generalisation and Transfer: E9--E11}
\paragraph{E9: Cross-age fall detection.}
Probes are trained on young-adult fall windows and evaluated on elderly subjects, measuring whether fall kinematics transfer across age groups. This is a strict train/test split rather than cross-validation.

\paragraph{E10: Fall direction classification.}
This probe tests whether the latent space preserves directional geometry of falls, distinguishing forward, backward, and lateral events. Macro-F1 is the primary summary because all directions are clinically meaningful.

\paragraph{E11: Cross-dataset HAR transfer.}
A probe fit on UCI HAR is applied without retraining to PAMAP2 after harmonizing the shared activity label set. This evaluates whether activity-class structure persists across datasets, devices, and recording conditions.

\subsection{Representation Quality and Structure: E12--E14}
\paragraph{E12: Post-hoc physical decoder.}
A shallow decoder attempts to recover coarse physical observables from frozen embeddings, including mean acceleration and RMS angular velocity. In the current table we report validation MAE directly, reserving any normalization by a persistence baseline for a later summary pass.

\paragraph{E13: Layer-wise effective rank.}
Activation matrices extracted from intermediate layers are analyzed for effective rank and complemented by latent-dynamics separation and latent-trajectory straightness diagnostics. Taken together, these probes are intended to detect collapse, bottlenecking, or overcompression across the depth and dynamics of the encoder.

\paragraph{E14: Horizon forecast.}
Frozen representations are used as features for supervised forecasting at $h{=}48$. Forecast MAE indicates how much short-term dynamical information survives in the learned state; broader multi-horizon summaries can be restored once the full sweep is complete.

\subsection{How the Main Text Uses the Battery}
The main text does not need to report every probe at equal depth. Section~\ref{sec:results} foregrounds two controlled comparisons: the principal shared-backbone \textsc{LIN} versus \textsc{MAE} objective comparison (Section~\ref{subsec:objective_comparison}), and a predictor-head ablation across four predictor classes of increasing expressivity (Section~\ref{subsec:predictor_ablation}). The appendix then serves as the exhaustive reference for what each probe means, why it matters, and how to interpret wins, losses, or mixed results.

\section{Mathematical Notation}
\label{app:notation}

\begin{table}[h]
  \centering
  \small
  \caption{Production configuration of \sonata.}
  \label{tab:config}
  \begin{tabular}{llr}
    \toprule
    Parameter & Symbol & Value \\
    \midrule
    Context / signal length & $L\,/\,L_\text{data}$  & $512\,/\,464$ \\
    Forecast horizon         & $p$                    & 48 \\
    Input channels           & $C$                    & 6 \\
    Embedding dim            & $d$                    & 128 \\
    FFN hidden dim           & $d_\text{ffn}$         & 768 \\
    GDN heads                & $H$                    & 4 \\
    Head / value dim         & $d_k\,/\,d_v$          & $32\,/\,256$ \\
    Layer stack              & ---                    & \texttt{[C,C,C,G]\textsuperscript{x3}} \\
    Parameters               & ---                    & ${\approx}3.77\text{\,M}$ \\
    \bottomrule
  \end{tabular}
\end{table}

\begin{table}[h]
  \centering
  \small
  \caption{Primary mathematical notation used in the \sonata architecture and objective formulation.}
  \label{tab:symbols}
  \begin{tabular}{lp{6.5cm}}
    \toprule
    \textbf{Symbol} & \textbf{Description} \\
    \midrule
    \multicolumn{2}{l}{\textit{Dimensions \& Indices}} \\
    $B, L, d$ & Batch size, sequence length ($512$), embedding dimension \\
    $p$ & Forecast horizon ($48$ steps) \\
    $C$ & Input/output signal channels (e.g., $6$ for IMU) \\
    \midrule
    \multicolumn{2}{l}{\textit{Data \& Representations}} \\
    $\mathbf{X}_t$ & Raw sensor context window at time $t$ \\
    $\mathbf{H}^{(j)}$ & Hidden representation at the $j$-th backbone layer \\
    $\mathbf{S}_t$ & Recurrent associative state / latent state \\
    $\hat{\mathbf{S}}_{t+1}$ & Predicted target latent state for the next window \\
    $\hat{\mathbf{Y}}$ & Kinematic forecast output \\
    \midrule
    \multicolumn{2}{l}{\textit{Architectural Components}} \\
    $\mathbf{q}_t, \mathbf{k}_t, \mathbf{v}_t$ & Queries, normalized keys, and values (GDN) \\
    $g_t, \beta_t$ & Content-dependent decay gate and update rate (GDN) \\
    $\pi_\phi$ & Expand--compress MLP projector \\
    $W_p$ & Lightweight linear predictor \\
    \midrule
    \multicolumn{2}{l}{\textit{Objective Function}} \\
    $\mathcal{L}_\text{LWM}$ & Total Latent World Model objective \\
    $\mathcal{L}_\text{pred}$ & Mean squared error prediction loss \\
    $\mathcal{L}_\text{SIGReg}$ & Sketched Isotropic Gaussian Regularization penalty \\
    $\lambda$ & Regularization weight trade-off \\
    \bottomrule
  \end{tabular}
\end{table}

\end{document}